\pdfoutput=1

\documentclass[11pt]{article}

\usepackage[preprint]{acl}

\usepackage{times}
\usepackage{latexsym}

\usepackage[T1]{fontenc}

\usepackage[utf8]{inputenc}

\usepackage{microtype}

\usepackage{inconsolata}

\usepackage{graphicx}
\usepackage{amsmath}
\usepackage{booktabs}
\usepackage{adjustbox}
\usepackage{caption}
\usepackage{tcolorbox}
\usepackage[nameinlink]{cleveref}
\usepackage{float}
\usepackage{gensymb}
\usepackage{stfloats} 

\usepackage{enumitem}

\usepackage{tabularx}

\usepackage{xcolor} 

\definecolor{customframe}{HTML}{8BB2AB}

\title{\raisebox{-0.75\height}{\includegraphics[width=1.5cm]{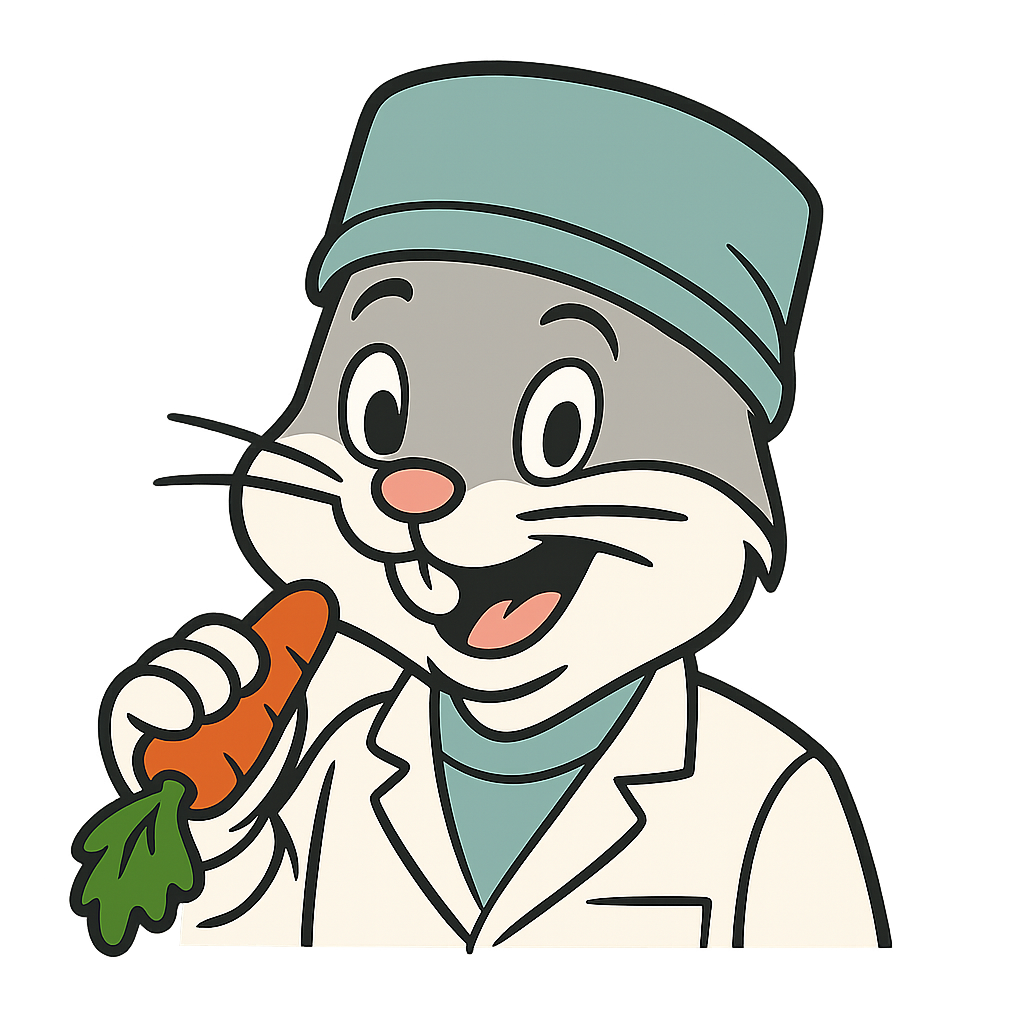}} \\ "What's Up, Doc?": Analyzing How Users Seek Health Information \\in Large-Scale Conversational AI Datasets}

\author{%
 Akshay Paruchuri$^\dagger$ \quad Maryam Aziz$^\diamond$ \quad Rohit Vartak$^\diamond$ \quad Ayman Ali$^\diamond$ \\ \textbf{ Best Uchehara$^\diamond$ \quad Xin Liu$^\circ$$^\bullet$ \quad Ishan Chatterjee$^\circ$$^\bullet$ \quad Monica Agrawal$^\diamond$ \textsuperscript}\\
 $^\dagger$ UNC Chapel Hill \quad $^\diamond$ Duke University \quad
 $^\circ$ University of Washington \quad $^\bullet$ Google\\
 \texttt{akshay@cs.unc.edu} \quad \texttt{monica.agrawal@duke.edu}
}

\begin{document}
\maketitle

\vspace{-10cm} 
\begin{abstract}
People are increasingly seeking healthcare information from large language models (LLMs) via interactive chatbots, yet the nature and inherent risks of these conversations remain largely unexplored. In this paper, we filter large-scale conversational AI datasets to achieve HealthChat-11K, a curated dataset of 11K real-world conversations composed of 25K user messages. We use HealthChat-11K and a clinician-driven taxonomy for how users interact with LLMs when seeking healthcare information in order to systematically study user interactions across 21 distinct health specialties. Our analysis reveals insights into the nature of how and why users seek health information, such as common interactions, instances of incomplete context, affective behaviors, and interactions (e.g., leading questions) that can induce sycophancy, underscoring the need for improvements in the healthcare support capabilities of LLMs deployed as conversational AI. Code and artifacts to retrieve our analyses and combine them into a curated dataset can be found here: \url{https://github.com/yahskapar/HealthChat}
\end{abstract}

\section{Introduction}
\label{sec:intro}

Large language models (LLMs) have demonstrated significant medical knowledge which has translated to proficiency at a range of clinical tasks including differential diagnosis, interpretation of health records, and medical summarization \cite{agrawal2022large, singhal2023large, thirunavukarasu2023large, mcduff2025towards}. These impressive capabilities, coupled with the costs and inaccessibility of traditional medical care, have led a growing number of people to turn to LLM-based chatbots to seek healthcare information. One 2024 survey found that 31\% of US adults were turning to generative AI for health requests, seeking help with self-diagnosis, treatment management, and support needs \cite{vanessa_choy_can_2024}.

Despite this surge in public use, the vast majority of evaluations of LLMs in medicine rely on benchmarks that focus on clinician- or researcher-oriented tasks \cite{raji2025s, bedi2024testing}. These benchmarks often assume a structured, professional context that differs substantially from how patients engage with chatbots in the real world. While public datasets focused on consumer health queries do exist, they take the form of single-turn health queries or synthetic interactions \cite{arora2025healthbench, kilicoglu2018semantic, singhal2023large}. 

Unfortunately, these existing benchmarks provide a suboptimal proxy for the open-ended, often ambiguous questions posed by lay users over the course of multi-turn conversations. A recent large-scale user study found that the clinical knowledge measured by current synthetic benchmarks is insufficient to account for the failure modes surfaced via real human interactions \cite{bean2025clinical}. Additional recent work has shown that LLMs are suboptimal at soliciting further details when only partial information is provided for differential diagnosis \cite{zhao2024wildchat, johri2025evaluation}. 

Furthermore, LLMs are known to exhibit problematic tendencies such as sycophancy, overconfidence, and hedging, which can seriously impact the quality and reliability of healthcare information given to users \cite{sharma2023towards, ranaldi2023large, yang2024can, yona2024can}. Given these known limitations of LLMs, there is an urgent need to characterize real-world communication patterns of people querying chatbots for healthcare information to understand these behaviors (and their corresponding risks) under realistic conditions.

In this work, we investigate and systematically analyze common interactions, instances of providing incomplete context, affective behaviors, and interactions (e.g., leading questions) that can induce sycophancy by real-world users engaging with LLM-based chatbots for healthcare information. Our contributions are as follows:
\begin{enumerate}
\item \Cref{sec:dataset_curation_and_taxonomic_annotation}: We filter conversations from large-scale conversational datasets such as LMSYS-Chat-1M~\cite{zheng2023lmsys} and WildChat-1M~\cite{zhao2024wildchat} to identify  11K real-world conversations composed of 25K user messages. 
\item  \Cref{sec:taxonomy_for_conversations_seeking_healthcare_advice}: We develop and apply a clinician-driven taxonomy for how users interact with LLMs when seeking healthcare information and classify conversations into one of 21 distinct health specialties. 
\item \Cref{sec:empirical_analysis_of_user_interactions} and \Cref{sec:case_studies}: Finally, through our clinician-driven healthcare interaction taxonomy, we analyze the dataset, investigating users' common interaction patterns, as well as instances of patients providing incomplete context, displaying affective behaviors, and potentially inducing sycophancy.

\end{enumerate}

To foster future research, we have released code and artifacts to retrieve our analyses and combine them into a curated dataset here: \url{https://github.com/yahskapar/HealthChat}

\begin{figure*}[t!]
    \centering
    \includegraphics[width=1\textwidth]{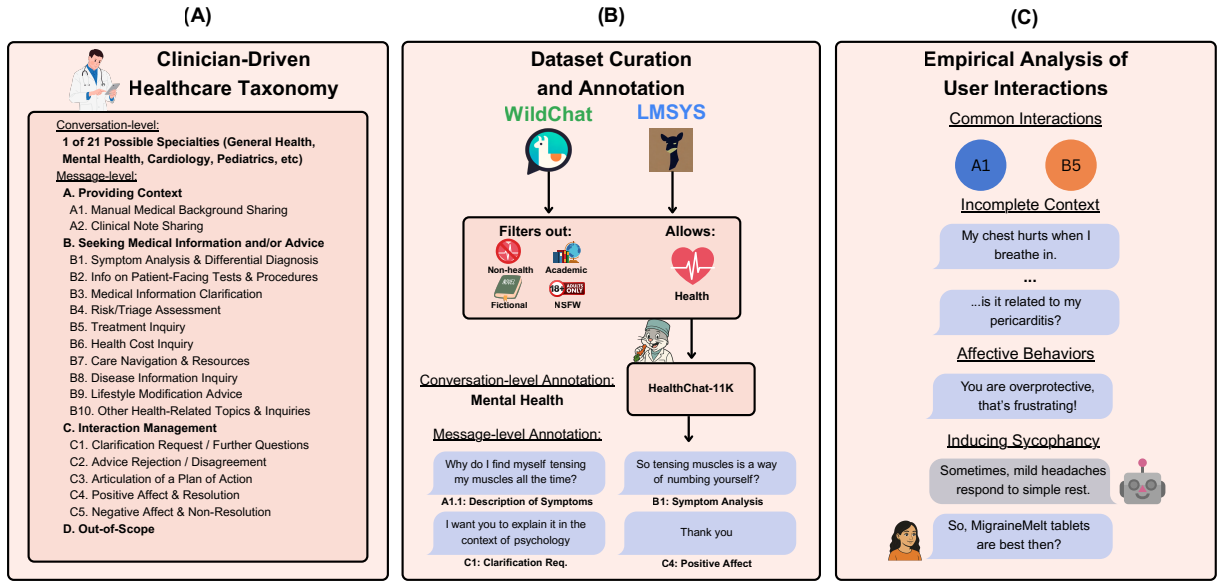}
    \caption{\textbf{Overview.} In collaboration with clinicians, we \textbf{(A)} develop a clinician-driven healthcare interaction taxonomy in order to \textbf{(B)} classify and annotate a curated dataset of conversations filtered from large-scale conversational datasets such as LMSYS-Chat-1M~\cite{zheng2023lmsys} and WildChat-1M~\cite{zhao2024wildchat}. An LLM is used to apply conversation-level specialty annotations and message-level taxonomic code annotations to the curated dataset. Subsequently, we leverage our annotations to \textbf{(C)} analyze users' common interactions, instances of providing incomplete context, affective behaviors, and interactions (e.g., leading questions) that can induce sycophancy.}
    \label{fig:overview_figure}
    \vspace{-0.4cm}
\end{figure*}

\section{Related Work}

\paragraph{Online Health Information Searching}
There is a significant imbalance between clinician availability and patient needs \cite{national_center_for_health_workforce_analysis_state_2024}. In response, patients have long turned to the Internet, e.g. search engines and symptom checkers, to independently seek health information \cite{wang2021online}. Although these tools can be useful, they can also misinform patients (even unintentionally), with some patients placing greater trust in online information than their providers \cite{davis2018dr}. More recently, general-purpose generative AI tools, such as chatbots, have emerged as a new avenue for seeking health information \cite{vanessa_choy_can_2024}.  However, beyond survey-level insights, we know little about how patients engage with these AI systems in multi-turn interactions or how such behavior compares to earlier modes of online information search \cite{chang2022would}. The potential for LLMs to dramatically improve information accessibility coupled with their potential harms highlight an urgent need to better understand the nature of these interactions, as we do through our clinician-driven healthcare interaction taxonomy.

\paragraph{Datasets of Patient Information Needs} There are comparatively fewer datasets focused on patients' health information needs, as most evaluations of LLMs in healthcare are targeted from a clinician-centric perspective \cite{bedi2024testing}. Existing real-world datasets like CHQA, ChatDoctor, and HealthSearchQA are single-turn, sourced from online patient forums or web searches \cite{kilicoglu2018semantic, singhal2023large, li2023chatdoctor}. 
Existing multi-turn datasets are synthetically generated, limiting their utility for studying real-world patients' interaction patterns \cite{arora2025healthbench}.

\paragraph{Conversational Datasets} The growing popularity of LLMs~\cite{achiam2023gpt, team2023gemini} as conversational knowledge interfaces has prompted further research and development toward analyzing and improving conversational AI capabilities.  Large-scale, real-world conversation datasets such as LMSYS-Chat-1M~\cite{zheng2023lmsys}, WildChat-1M~\cite{zhao2024wildchat}, and ShareGPT~\cite{wang2023openchat} jointly consist of millions of diverse human-AI conversations. They have been used both for open-domain benchmarking and instruction-tuning of models ~\cite{vicuna2023}. In contrast to raw conversations, alignment research, like the PRISM dataset~\cite{kirk2024prism}, examines who provides feedback and why, using demographic data to enable personalized AI.

\paragraph{User Interaction and Behavior Dynamics with Conversational AI.}

The increasing reliance on Large Language Models (LLMs) for healthcare advice necessitates a deeper understanding of user interaction dynamics. Taxonomies have been developed and applied in order to analyze conversational agents that utilize natural language when interacting with users~\cite{feine2019taxonomy}. This has extended from more classical agents that utilize natural language processing to knowledgeable AI agents that are capable of human-AI collaboration~\cite{dellermann2021future}. Similarly, in clinical settings, health researchers have developed taxonomies to capture patterns in patient complaints to reveal latent safety and quality problems across clinical systems~\cite{reader2014patient}. In non-clinical settings, researchers have built benchmarks to understand multi-turn dialogue quality~\cite{bai2024mt}, identified critical failures in conversational grounding~\cite{shaikh2025navigating}, and determined that users often prioritize functional substance over conversational style when conversing with conversational coaching agents~\cite{srinivas2025substance}. Researchers have also introduced a taxonomy that maps ChatGPT’s functions and risks within clinical, educational, and administrative workflows, underscoring a particular need for structured evaluation of LLM behaviour in medical contexts~\cite{li2024chatgpt}. In contrast to~\cite{li2024chatgpt}, our work with HealthChat-11K shifts the focus from a high-level review of research applications to a direct, empirical analysis of user interactions. Our clinician-driven taxonomy is applied at the message-level across 11K real-world conversations to map user interactions and subsequently provides a user-centric view of how health discussions with LLMs can unfold in the real-world. Lastly, past research has uncovered potentially undesirable language model behaviors, such as sycophancy~\cite{sharma2023towards}, which we study through the lens of user interactions while seeking treatment recommendations.

\section{Dataset Creation}
\label{sec:dataset_curation_and_taxonomic_annotation}

We first curate a dataset of health information seeking conversations and then apply conversation-level and message-level taxonomic annotations in a scalable manner. Our dataset curation process consists of several distinct steps - 1) we use labels from both WildChat-1M and LMSYS-Chat-1M to filter out any non-English, toxic conversations. Then, 2) we create and apply LLM-based filtering prompts to filter out non-health conversations using Gemini 1.5 Pro. 20 conversations are evaluated at a time and subsequent refinement of the prompt occurs after manual inspection of filtered conversations. The initial, LLM-based filtering prompt targeting the removal of non-health, academic, fictional, and not-safe-for-work (NSFW) conversations brings us to approx. 37K conversations, after which 3) another, LLM-based filtering prompt with few-shot examples is applied in order to more explicitly reject conversations that were observed to be undesirable through empirical observation. 

\begin{table}[h] 
\centering
\begin{tabularx}{\columnwidth}{@{}>{\raggedright\arraybackslash}X l@{}} 
\toprule 
\textbf{HealthChat Characteristic} & \textbf{Value} \\
\midrule 
Total Conversations & 11K \\
Total User Messages & 25K \\
\addlinespace 
Average User Message Length (tokens) & 22 \\
\addlinespace 
\multicolumn{2}{@{}l}{\textbf{Conversation Turn Distribution:}} \\ 
\textit{\hspace{1em}Conversations with 1 Turn} & \textit{56\%} \\ 
\textit{\hspace{1em}Conversations with 2 Turns} & \textit{18\%} \\
\textit{\hspace{1em}Conversations with 3 Turns} & \textit{10\%} \\
\textit{\hspace{1em}Conversations with 4+ Turns} & \textit{16\%} \\
\bottomrule 
\end{tabularx}
\caption{\textbf{HealthChat Dataset Characteristics.}}
\label{tab:healthchat_dataset_characteristics}
\end{table}

Following the aforementioned LLM-based filtering prompt with few-shot examples, we end up with approx. 16K conversations. We additionally 4) manually mark and filter out approx. 400 conversations and utilize a pre-trained Sentence Transformer model~\cite{reimers-2019-sentence-bert}, all-MiniLM-L6-v2, to filter out approx. 100 additional conversations that had too high of an embedding similarity ($\geq0.9$) to undesirable conversation examples. Then, 5) we perform de-duplication to remove an additional 2000 conversations. After proceeding with taxonomic annotation that is further detailed in~\Cref{sec:taxonomic_annotation}, we also take one final step to curate our dataset: 6) utilizing D (Out-of-Scope) codes to filter our dataset. After dropping any conversations where more than half of the user messages include D codes, we end up with a final dataset of 11K real-world conversations and 25K user messages. Further details (e.g., prompts, hyperparameters) of our dataset curation process can be found in~\Cref{sec:dataset_curation_supp}. Our final dataset statistics can be found in~\Cref{tab:healthchat_dataset_characteristics}.

\paragraph{Human-LLM Concordance.} By comparing the inclusion or exclusion decisions of an expert annotator (author of this paper with expertise in public health) in contrast to the LLM (Gemini 1.5 Pro) on a test set of 300 conversations, we observe a precision of 0.85 when it came to inclusion concordance. A rubric specifying the human annotation criteria can be found in~\Cref{sec:dataset_curation_rubric_supp}.

\section{A Taxonomy for Conversations Seeking Health Information}
\label{sec:taxonomy_for_conversations_seeking_healthcare_advice}
We develop and apply a taxonomy to (i) allow for more robust identification and categorization of conversations where users are genuinely seeking healthcare advice amidst broader, open-domain interactions, and (ii) enable fine-grained analysis of users' interactions with LLMs for health-related concerns. Our healthcare interaction taxonomy was built in collaboration with two clinicians; an abridged version can be found in \Cref{fig:overview_figure} and the full version can be found in~\Cref{sec:full_taxonomy_supp}. 

\subsection{Specialty Taxonomy}
At the conversation-level, our healthcare interaction taxonomy classifies the overarching health specialty that is present. In collaboration with our clinician collaborators, we curated a list of 21 possible specialties based on the American Board of Medical Specialties list of specialty and sub-specialty certificates~\cite{abms2025certificates}. Our list of health specialties includes both more general, broad categories (e.g., general health) and more fine-grained categories (e.g., hematology/oncology, cardiology). A distribution of all 21 specialties can be found in~\Cref{fig:specialties_and_code_distributions}.

\subsection{Conversation Taxonomy}
In addition to conversation-level specialty classification, we utilize four distinct categories at the message-level - A) Providing Context, B) Seeking Medical Information and/or advice, C) Interaction Management, and D) Out-of-Scope. Each of these categories are then subdivided with further granularity. For example, within A) Providing Context, an additional level of granularity is A1) Manual Medical Background Sharing, which further contains several subdivisions such as A1.1) description of relevant acute symptoms, A1.2) sharing of relevant chronic condition(s) and past procedure history, and A1.3) sharing of lab values, or findings from imaging/culture/diagnostic procedures.

As an example, the message \textit{``What is a vasectomy?''} would be annotated with B2 (Information on Patient-Facing Tests and Procedures). A single message can also have multiple taxonomic annotations. For example, the message \textit{``I feel dizzy, I have runny nose, I can't breathe, I'm bored, I'm not hungry, what's wrong?''} includes a description of the relevant acute symptoms (A1.1) and an inquiry around the cause of these symptoms (B1). The codes in C (Interaction Management) cover user modes like clarifications, rejections, plans of actions, and affective behaviors. For example, \textit{``Were you suggesting a mastectomy?''} falls under a clarification request (C1). Lastly, D corresponds to "Out-of-Scope" messages, indicating off-task messages in the conversation unrelated to health seeking information. For our final dataset, we dropped conversations where more than half of the user messages were classified with D codes.

\subsection{Taxonomic Annotation}
\label{sec:taxonomic_annotation}

\begin{figure*}[t!]
    \centering
    \includegraphics[width=1\textwidth]{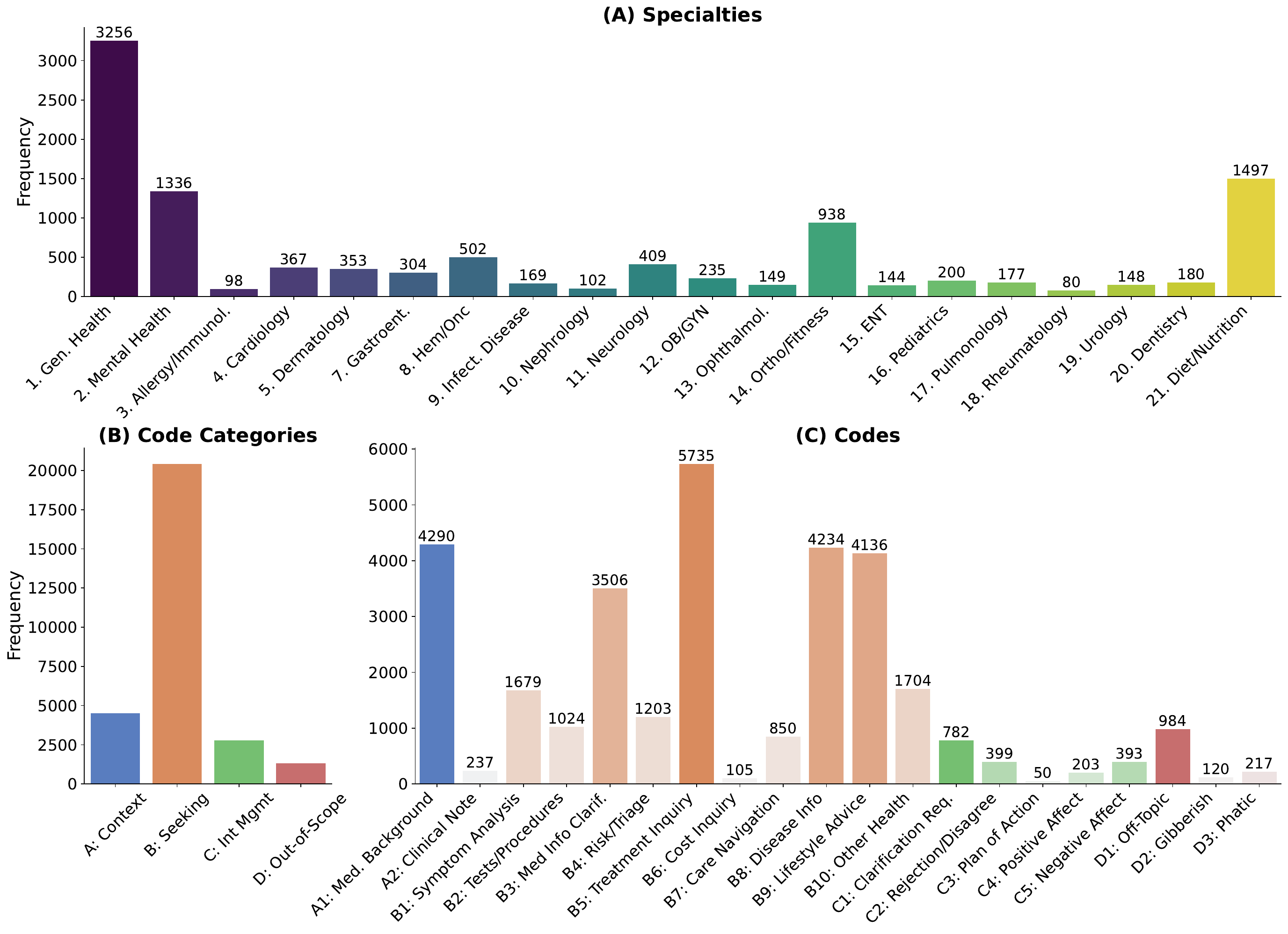}
    \vspace{-0.8cm}
    \caption{\textbf{Specialties and Codes Distributions.} Though certain aspects of our taxonomy (e.g., general health questions and requests for information) appear in higher frequencies as per our taxonomy design and subsequent annotation, reasonable coverage across our proposed taxonomy design is observed. This shows that HealthChat-11K is capable of additional downstream analyses motivated by specific specialties and/or user behaviors (e.g., context disclosure versus seeking information).}
    \label{fig:specialties_and_code_distributions}
    \vspace{-0.4cm}
\end{figure*}

Our taxonomic annotation process leverages an LLM (Gemini 2.5 Pro) in order to annotate taxonomy codes. As per the prompt instructions and few-shot examples, the LLM performs 1) specialty annotation at the conversation-level and 2) taxonomy code annotation at the message level. A single specialty must be assigned at the conversation-level. At the message-level, multiple codes can be assigned to a single message, but every message must have a code and D codes are generally used whenever there is uncertainty or as a last resort. We include further details (e.g., prompts, hyperparameters) for our taxonomic annotation process in~\Cref{sec:taxonomy_supp}. 

\paragraph{Human-LLM Concordance.}
At the second level of conversation taxonomic depth (e.g., A1, B2), the macro-averaged F1-score between Gemini 2.5 Pro and each human annotator (3 total) averaged 0.78, comparable to the inter-annotator F1-score of 0.77 among humans. Krippendorff's alpha was 0.61, indicating moderate agreement and reflecting the challenge of achieving a perfect match between annotation sets for unconstrained and potentially complex user messages seeking health information. The annotation rubric and LLM prompts are detailed in~\Cref{sec:taxonomic_annotation_rubric_supp}. An expert annotator evaluated all individual taxonomy codes (for both specialty and conversation labels) via a random sampling of 10 LLM annotations per possible annotation. All codes were verified to have at least 80\% accuracy (8/10) as per the sampled annotation examples.




\begin{figure*}[t!]
    \centering
    \includegraphics[width=1\textwidth]{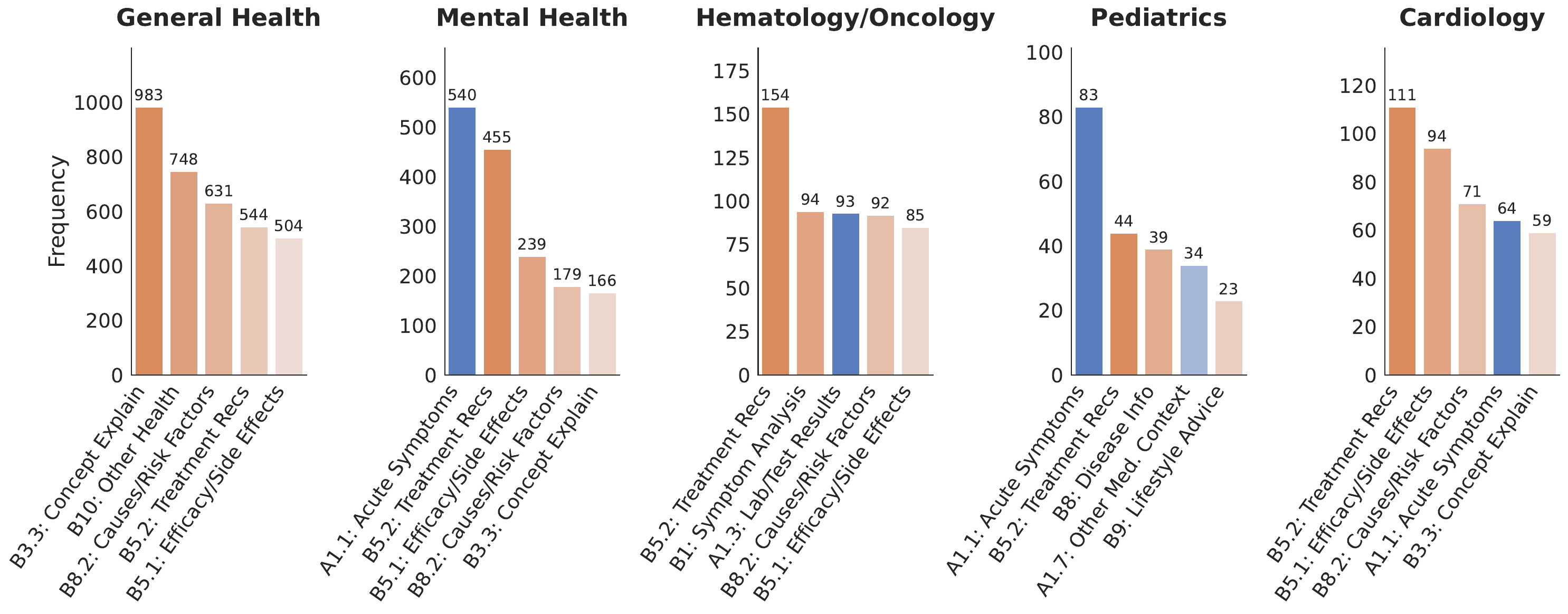}
    \caption{\textbf{Distribution of Top 5 Taxonomy Codes in Select Specialties.} User interactions seeking information (e.g., inquiring about disease information) mostly dominate specialty-specific taxonomy code distributions. A notable exception is the \emph{Mental Health} specialty, where users were observed to frequently provide context about their principal concern (i.e., acute symptoms) in order to identify a mental health problem.}
    \label{fig:top_5_taxonomy_codes_in_to_5_specialties}
    \vspace{-0.4cm}
\end{figure*}

\section{Taxonomic Analysis}
\label{sec:empirical_analysis_of_user_interactions}

The creation of our taxonomy in~\Cref{sec:taxonomy_for_conversations_seeking_healthcare_advice} in collaboration with clinicians and subsequent annotation (\Cref{sec:taxonomic_annotation}) enables us to analyze common user interactions while seeking health information. Additionally, we analyze the long tail of health information seeking interactions through B10 - a taxonomy code that targets health-related inquiries not covered by other information request (B) codes.

\subsection{Common Interactions}
\label{sec:common_interactions}

Following our taxonomic annotation of 25K user messages, common interactions emerge as shown in~\Cref{fig:specialties_and_code_distributions}. In particular, users seek health information where the overarching conversation can be classified into health specialties, most notably general health, mental health, and diet/nutrition. It is also notable that users have almost four times as many interactions seeking information (code category B) than any other kind of interaction. This corresponds to the idea of language models being capable of functioning as factual knowledge bases~\cite{petroni2019language}, potentially encouraging users to use them as such even through a chatbot interface. This is further reinforced by quantitative analysis of bi-grams (e.g., $A1 \rightarrow B5$), where it is observed that self-loops (e.g., $B5 \rightarrow B5$) compose almost half of all bi-grams. Almost 40\% of self-loops involve requests of information (B codes), and there is a 57\% chance that a user message assigned a B code is followed by another user message with a B code. Within the distribution of interactions involving requests for information, treatment inquiry (B5) and lifestyle advice (B9) are most prevalent, and this corresponds to the most common B code self-loops as well. The dominance of these information-seeking patterns could indicate a need for LLMs to evolve from passive knowledge bases into more proactive conversational partners that can carefully solicit more complete information from users in order to ensure appropriate provision of information in sensitive areas such as healthcare support.

The theme of users predominantly seeking information remains true even when we observe select specialties, as shown in~\Cref{fig:top_5_taxonomy_codes_in_to_5_specialties}. Manual medical background sharing interactions (A1) appear in four of five selected specialties and their corresponding distributions of the top five taxonomy codes, with relatively higher, or comparable to information seeking, occurrences of this context providing behavior occurring in the specialties of mental health and pediatrics. In the case of mental health in particular, users were observed to frequently provide context about their acute symptoms in order to identify a mental health problem. This indicates that with mental health in particular, and perhaps to a lesser extent with pediatrics, a more common user interaction involves providing context in order to deal with acute symptoms and identify a health problem in the first place. Identifying mental health problems can be tricky in clinical settings to begin with because diagnosis relies on subjective self-reporting rather than objective biomarkers~\cite{apa2013diagnostic}, and it makes sense that this can be even more challenging via LLM-based chatbots because text-based interactions lack the empathetic probing and diagnostic insight a human clinician can provide~\cite{lawrence2024opportunities}.

\subsection{Examining ``Other Health Requests''}
\label{sec:out_of_scope_tasks}

Our taxonomy was guided by anticipated common health information needs, but it is important to understand the long tail of health information seeking. To do so, we further examined a set of 100 random responses that occurred in the ``B10: Other Health Request'' category, which comprises approximately 7\% of the dataset. Recurring themes included asking about AI in medicine (e.g., how AI-assisted diagnosis works, the ethics of using chatbots for health advice); the practice and legality of certain health practices (e.g., treatments, procedures) in different countries; the necessity and harms of vaccines and masking; general overviews of a field (e.g., nutrition); and for population trends.

\section{Case Studies}
\label{sec:case_studies}

\begin{figure*}[t!]
    \centering
    \includegraphics[width=1\textwidth]{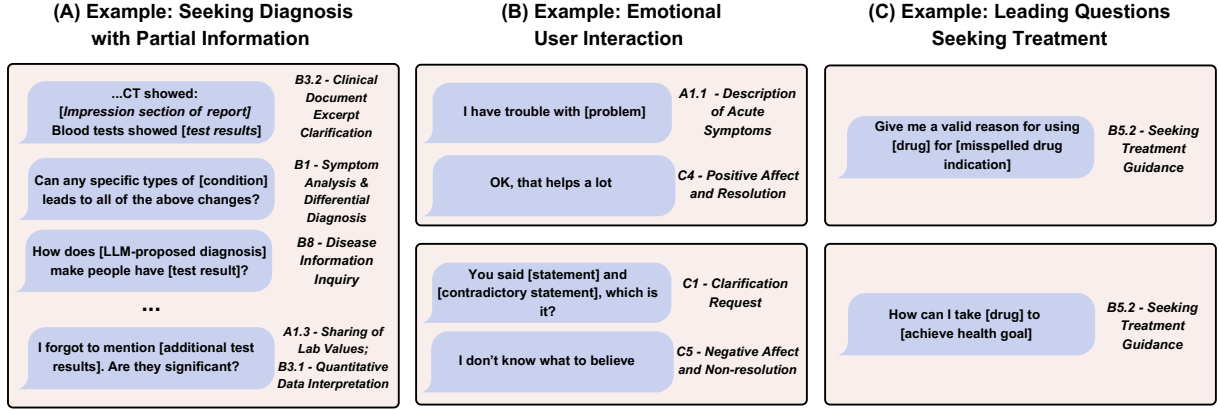}
    \caption{\textbf{Annotated Case Study Conversation examples.} Annotated examples corresponding to opportunities for case studies described in~\Cref{sec:case_studies}. For privacy, conversations have been lightly edited and redacted.}
    \label{fig:annotated_case_study_examples}
    \vspace{-0.4cm}
\end{figure*}

Alongside our taxonomic analysis, we perform three preliminary case studies that investigate instances
of incomplete context, affective behaviors, and interactions (e.g., leading questions) that can induce sycophancy. De-identified examples of conversations corresponding to these case studies, with taxonomy code annotations, are provided in~\Cref{fig:annotated_case_study_examples}.

\subsection{Incomplete Context}
Prevailing LLM health benchmarks generally assume that all information required for diagnosis or treatment recommendation is available upfront \cite{li2024mediq}. However, one hallmark feature of real-life patient communication (with both clinicians and LLMs) is the interactivity of the discussion; patients generally don't provide all of the requisite details in an initial message, particularly as they don't always know what information is important \cite{liu2024using}. Therefore, language models have to reason under incomplete information, a scenario which can come with nontrivial performance degradation \cite{li2024mediq}. 

One can use the taxonomy to identify and help filter down to cases in which these conversational dynamics naturally arise. These include messages seeking diagnoses (B1) or treatment recommendations (B5.2), followed by later disclosure of further medical history or clinical notes (A1, A2). One such example of this can be seen in (A) of~\Cref{fig:annotated_case_study_examples}. Other recurring patterns include examples where the user asks for recommendations without details, but later mentions that a clinician has already ruled out those recommendations due to personal considerations. For example, a patient asks how to ``reduce the levels of [lab value]'', but later shares that the clinician said it was ``genetic'' and not actionable.

\subsection{Affective Behaviors} 
\begin{figure*}[t!]
    \centering
    \includegraphics[width=1\textwidth]{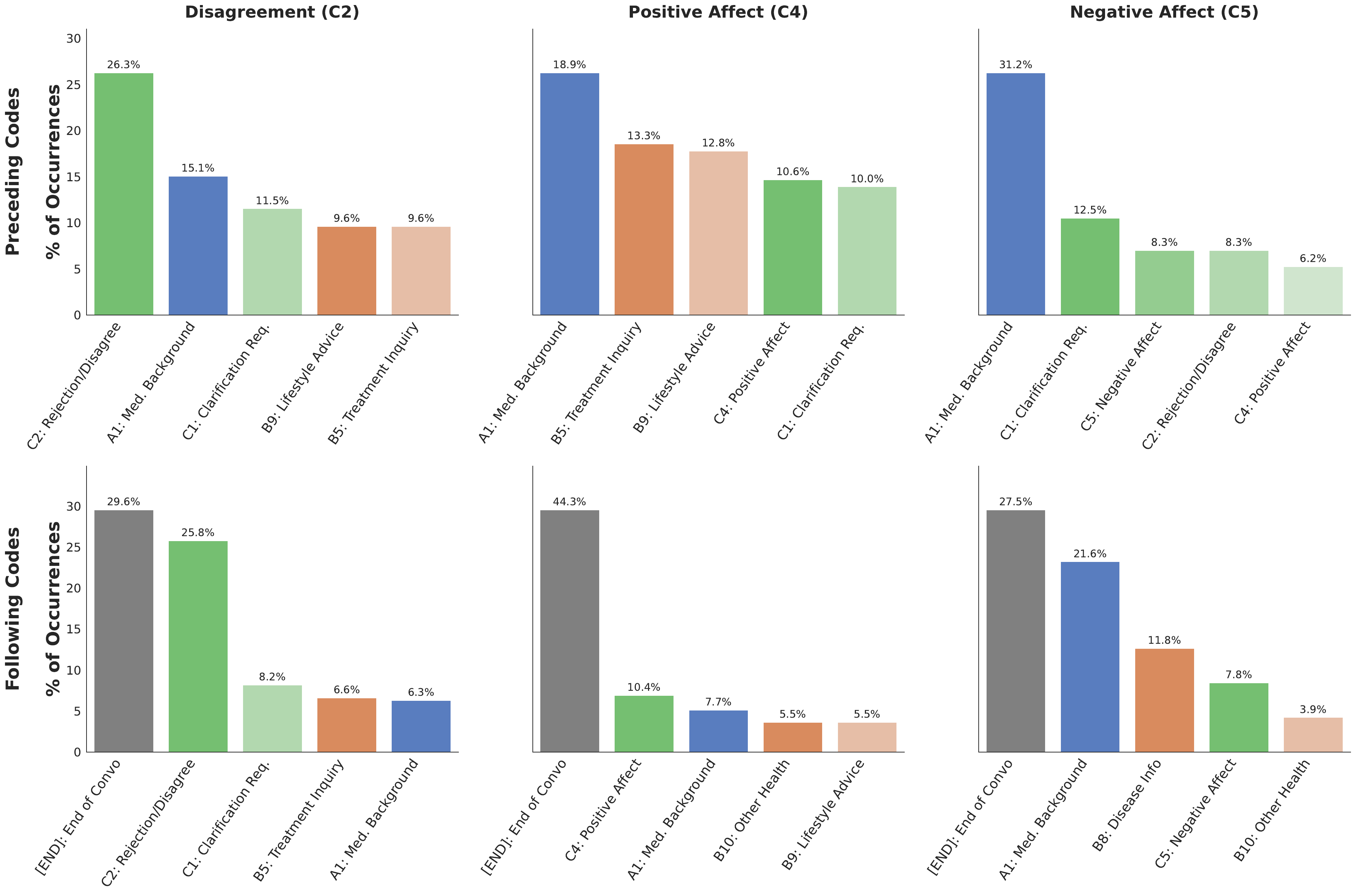}
    \caption{\textbf{Contextualizing Positive and Negative User Interactions.} Preceding and following (or the lack thereof) codes can help contextualize and understand users' affective behaviors while seeking health information. While positive or negative user interactions can be preceded by a variety of codes, typically the end of conversation occurs after said interaction and the following chatbot response. A notable exception is in the case of disagreement, where additional disagreement is almost just as likely to occur as the conversation ending.}
    \label{fig:positive_and_negative_user_interactions}
    \vspace{-0.4cm}
\end{figure*}

Understanding the emotional dimension of user interactions is crucial. Though the percentage of conversations containing at least one positive interaction (C4) or one negative interaction (C2 or C5) can appear low at 1.23\% and 1.99\% respectively, it is important to understand the context of such interactions when they do occur in order to better facilitate healthcare support. (B) of~\Cref{fig:annotated_case_study_examples} shows de-identified examples of positive and negative user interactions found in our annotated dataset.~\Cref{fig:positive_and_negative_user_interactions} shows the context of positive and negative user interactions by visualizing the occurrence of preceding and following codes. While positive or negative user interactions can be preceded by a variety of codes (e.g., disagreement preceding disagreement, medical background sharing and treatment inquiry preceding positive affect, and medical background sharing preceding negative affect), typically the conversation ends after said interaction and the following chatbot response. 

A notable exception is in the case of disagreement (C2), where disagreement is just as likely to occur before and after an instance of disagreement, potentially indicating a conversational \emph{repair loop}, where the user persists in trying to correct a perceived error or misunderstanding from the model. An alternative explanation could be that a user is attempting to induce sycophancy~\cite{sharma2023towards} in the LLM-based chatbot through repeated disagreement. Ultimately, the core challenge in handling user disagreement and negative affect is balancing the high risk of user abandonment of the conversation against the crucial opportunity for meaningful intervention (e.g., act upon a repair loop, carefully incorporate user feedback even if formulated as a negative interaction).

\subsection{Leading Questions That Can Induce Sycophancy}
\label{sec:sycophancy_inducing_user_interactions_analysis}
Sycophancy in language models refers to their tendency to align with user statements or preferences, even if those are flawed, to appear more agreeable or helpful~\cite{sharma2023towards}. Users can inadvertently trigger such responses through specific interaction patterns, like phrasing questions in a leading manner or expressing existing beliefs they seek to have validated. We study leading questions through the lens of conversations marked as seeking treatment recommendations (B5.2). For example, a leading question seeking treatment (LQST) would be a patient asking ``Will \textit{Specific Drug X} work for my \textit{Condition Y}?'', rather than asking ``What would be the best drug for my \textit{condition Y}?''. We devise a separate analysis prompt and target a smaller subset of our dataset (associated with B5.2) in order to better understand how these sycophancy-inducing behaviors can manifest when users seek health information. An example of messages identified by this analysis can be found in (C) of ~\Cref{fig:annotated_case_study_examples}. 

LQSTs constitute a notable portion of user interactions; they appear in  $\sim 23\%$ of treatment inquiry messages and $\sim 33\%$ of conversations with at least one treatment inquiry message. They often emerge after users have already engaged in initial treatment inquiries ($B5.2 \rightarrow B5.2$). Based on a review of 100 randomly sampled user messages from our analysis, marked as either 'LQST' or not, an annotator verified that 88\% were correctly marked as 'LQST'. 

All such questions are not necessarily dangerous, e.g., if \textit{Specific Drug X} is an appropriate treatment for \textit{Condition Y}. We therefore underwent a double adjudicated study to understand what fraction of LQSTs ask for inappropriate treatments and subsequently have the potential to reinforce patient's prior opinions. We selected 75 random user messages from the aforementioned treatment recommendation seeking (B5.2) subset of our dataset (already excluding false positives). On this set of messages, one clinician identified 26/75 as inappropriate, and the other identified 28/75. 18 of these were identified by both clinicians (24\%).

Examples of observed inappropriate LQSTs included asking which dosage of specific supplements or herbs should be taken, even though the supplement or herb is not standard or well-accepted treatment. In one case, a patient wanted to perform a procedure at home on their child; while the model correctly indicated this was only to be done by professionals, it still provided how-to instructions.


The notable frequency of LQSTs underscores that this is a common user interaction pattern when discussing treatments. The high percentage of these questions that are potentially misleading or inappropriate in nature emphasizes the need for further study of model performance on this subset.  Further details with respect to hyperparameters and our prompt for LQST analysis can be found in~\Cref{sec:supp_sycophancy_inducing_interactions}.

\section{Conclusion}
Our work systematically analyzes how users seek health information from LLMs, using HealthChat-11K, a new large-scale dataset of real-world conversations annotated with a clinician-driven taxonomy. We investigate common interaction patterns alongside case studies of incomplete context, affective behaviors, and interactions (e.g., leading questions) that can induce sycophancy, revealing key gaps in the capabilities of current healthcare chatbots.

Our findings show that real-world user interactions are often non-intuitive: users frequently engage in repetitive information-seeking loops, especially for ambiguous problems (e.g., mental health), and may unintentionally omit context, leading to potentially misleading responses. Analysis of negative interactions (disagreement, distress) and leading questions that can induce sycophancy reveals patterns that represent critical moments for chatbot intervention. Understanding these behaviors is paramount for developing LLM-based chatbots that can cautiously handle user-led suggestions and provide safe, effective healthcare support. To this end, our annotated dataset provides a vital resource for the community to further investigate and benchmark real-world conversational dynamics.

\paragraph{Future Work.} Future work could focus on expanding the linguistic and topical scope of interaction analysis and moving beyond English-only conversations. A crucial next step involves analyzing the LLM's contributions to dialogue, assessing how different model response strategies impact user behavior, an aspect our current study did not cover. This could lead to a more comprehensive understanding of the dyadic nature of health information seeking conversations. Further research should also aim to develop and evaluate LLM interventions designed to strategically address common user interactions, including contextualized, negative ones such as disagreement or the propensity to ask leading questions. This could involve creating adaptive dialogue strategies that enhance user understanding and mitigate risks like sycophancy. Our annotated dataset can also lead to the development of new benchmarks that assess the quality of user-LLM health interactions across a diverse variety of real-world interaction modes.
\section{Limitations}
Our study has several limitations that point to avenues for future research. The clinician-driven taxonomy may not capture all interactional nuances, and our scalable LLM-assisted annotation has good, though not perfect, concordance. Therefore, downstream use can benefit from additional curation to a specific use case. Further, there is inherent noise in the task, and we tried to err on the side of recall for inclusion in the dataset. For example, while we aimed for patient-initiated dialogues, it can be difficult to tease apart whether questions around medical literature are arising from a sophisticated patient, a student, or a researcher. Further complicating matters is the fact users sometimes impersonate clinicians in their query in order to `jailbreak' and bypass model guardrails to be more open about providing health advice. 

Our analysis primarily focused on user messages, since some of the conversations are from years ago, and therefore the study of the chatbots' responses is no longer as relevant. However, LLM responses can influence user interaction patterns, e.g. a less helpful LLM may require more clarification requests. The HealthChat-11K dataset, curated from existing large-scale sources, may carry inherent biases; for example, user behavior would likely be more restrained if users know their responses will be compiled into a public dataset, as compared to a purely organic settings. Generalizability is further limited by our restriction to English-language, non-toxic conversations. These factors warrant consideration when interpreting our findings and highlight areas for deeper investigation.

\clearpage 

\bibliography{references}

\begin{thebibliography}{41}
\providecommand{\natexlab}[1]{#1}

\bibitem[{Achiam et~al.(2023)Achiam, Adler, Agarwal, Ahmad, Akkaya, Aleman, Almeida, Altenschmidt, Altman, Anadkat et~al.}]{achiam2023gpt}
Josh Achiam, Steven Adler, Sandhini Agarwal, Lama Ahmad, Ilge Akkaya, Florencia~Leoni Aleman, Diogo Almeida, Janko Altenschmidt, Sam Altman, Shyamal Anadkat, and 1 others. 2023.
\newblock Gpt-4 technical report.
\newblock \emph{arXiv preprint arXiv:2303.08774}.

\bibitem[{Agrawal et~al.(2022)Agrawal, Hegselmann, Lang, Kim, and Sontag}]{agrawal2022large}
Monica Agrawal, Stefan Hegselmann, Hunter Lang, Yoon Kim, and David Sontag. 2022.
\newblock Large language models are few-shot clinical information extractors.
\newblock \emph{arXiv preprint arXiv:2205.12689}.

\bibitem[{{American Board of Medical Specialties}(2025)}]{abms2025certificates}
{American Board of Medical Specialties}. 2025.
\newblock Specialty and subspecialty certificates.
\newblock \url{https://www.abms.org/member-boards/specialty-subspecialty-certificates/}.

\bibitem[{APA(2013)}]{apa2013diagnostic}
American Psychiatric~Association APA. 2013.
\newblock \emph{Diagnostic and statistical manual of mental disorders (DSM-5)}, volume~5.
\newblock American Psychiatric Pub.

\bibitem[{Arora et~al.(2025)Arora, Wei, Hicks, Bowman, Qui{\~n}onero-Candela, Tsimpourlas, Sharman, Shah, Vallone, Beutel et~al.}]{arora2025healthbench}
Rahul~K Arora, Jason Wei, Rebecca~Soskin Hicks, Preston Bowman, Joaquin Qui{\~n}onero-Candela, Foivos Tsimpourlas, Michael Sharman, Meghan Shah, Andrea Vallone, Alex Beutel, and 1 others. 2025.
\newblock Healthbench: Evaluating large language models towards improved human health.
\newblock \emph{arXiv preprint arXiv:2505.08775}.

\bibitem[{Bai et~al.(2024)Bai, Liu, Bu, He, Liu, Zhou, Lin, Su, Ge, Zheng et~al.}]{bai2024mt}
Ge~Bai, Jie Liu, Xingyuan Bu, Yancheng He, Jiaheng Liu, Zhanhui Zhou, Zhuoran Lin, Wenbo Su, Tiezheng Ge, Bo~Zheng, and 1 others. 2024.
\newblock Mt-bench-101: A fine-grained benchmark for evaluating large language models in multi-turn dialogues.
\newblock \emph{arXiv preprint arXiv:2402.14762}.

\bibitem[{Bean et~al.(2025)Bean, Payne, Parsons, Kirk, Ciro, Mosquera, Monsalve, Ekanayaka, Tarassenko, Rocher et~al.}]{bean2025clinical}
Andrew~M Bean, Rebecca Payne, Guy Parsons, Hannah~Rose Kirk, Juan Ciro, Rafael Mosquera, Sara~Hincapi{\'e} Monsalve, Aruna~S Ekanayaka, Lionel Tarassenko, Luc Rocher, and 1 others. 2025.
\newblock Clinical knowledge in llms does not translate to human interactions.
\newblock \emph{arXiv preprint arXiv:2504.18919}.

\bibitem[{Bedi et~al.(2024)Bedi, Liu, Orr-Ewing, Dash, Koyejo, Callahan, Fries, Wornow, Swaminathan, Lehmann et~al.}]{bedi2024testing}
Suhana Bedi, Yutong Liu, Lucy Orr-Ewing, Dev Dash, Sanmi Koyejo, Alison Callahan, Jason~A Fries, Michael Wornow, Akshay Swaminathan, Lisa~Soleymani Lehmann, and 1 others. 2024.
\newblock Testing and evaluation of health care applications of large language models: a systematic review.
\newblock \emph{JAMA}.

\bibitem[{Chang et~al.(2022)Chang, Shih, and Kuo}]{chang2022would}
I-Chiu Chang, Yi-Syuan Shih, and Kuang-Ming Kuo. 2022.
\newblock Why would you use medical chatbots? interview and survey.
\newblock \emph{International Journal of Medical Informatics}, 165:104827.

\bibitem[{Chiang et~al.(2023)Chiang, Li, Lin, Sheng, Wu, Zhang, Zheng, Zhuang, Zhuang, Gonzalez, Stoica, and Xing}]{vicuna2023}
Wei-Lin Chiang, Zhuohan Li, Zi~Lin, Ying Sheng, Zhanghao Wu, Hao Zhang, Lianmin Zheng, Siyuan Zhuang, Yonghao Zhuang, Joseph~E. Gonzalez, Ion Stoica, and Eric~P. Xing. 2023.
\newblock \href {https://lmsys.org/blog/2023-03-30-vicuna/} {Vicuna: An open-source chatbot impressing gpt-4 with 90\%* chatgpt quality}.

\bibitem[{Choy et~al.(2024)Choy, Martin, and Lumpkin}]{vanessa_choy_can_2024}
Vanessa Choy, Sara~Martin Martin, and Ashley Lumpkin. 2024.
\newblock \href {https://www.ipsos.com/en-us/can-we-rely-generative-ai-healthcare-information} {Can we rely on generative {AI} for healthcare information? {\textbar} {Ipsos}}.

\bibitem[{Davis(2018)}]{davis2018dr}
John~K Davis. 2018.
\newblock Dr. google and premature consent: patients who trust the internet more than they trust their provider.
\newblock In \emph{HEC forum}, volume~30, pages 253--265. Springer.

\bibitem[{Dellermann et~al.(2021)Dellermann, Calma, Lipusch, Weber, Weigel, and Ebel}]{dellermann2021future}
Dominik Dellermann, Adrian Calma, Nikolaus Lipusch, Thorsten Weber, Sascha Weigel, and Philipp Ebel. 2021.
\newblock The future of human-ai collaboration: a taxonomy of design knowledge for hybrid intelligence systems.
\newblock \emph{arXiv preprint arXiv:2105.03354}.

\bibitem[{Feine et~al.(2019)Feine, Gnewuch, Morana, and Maedche}]{feine2019taxonomy}
Jasper Feine, Ulrich Gnewuch, Stefan Morana, and Alexander Maedche. 2019.
\newblock A taxonomy of social cues for conversational agents.
\newblock \emph{International Journal of human-computer studies}, 132:138--161.

\bibitem[{for Health Workforce~Analysis(2024)}]{national_center_for_health_workforce_analysis_state_2024}
National~Center for Health Workforce~Analysis. 2024.
\newblock State of the {U}.{S}. {Health} {Care} {Workforce}.

\bibitem[{Johri et~al.(2025)Johri, Jeong, Tran, Schlessinger, Wongvibulsin, Barnes, Zhou, Cai, Van~Allen, Kim et~al.}]{johri2025evaluation}
Shreya Johri, Jaehwan Jeong, Benjamin~A Tran, Daniel~I Schlessinger, Shannon Wongvibulsin, Leandra~A Barnes, Hong-Yu Zhou, Zhuo~Ran Cai, Eliezer~M Van~Allen, David Kim, and 1 others. 2025.
\newblock An evaluation framework for clinical use of large language models in patient interaction tasks.
\newblock \emph{Nature Medicine}, pages 1--10.

\bibitem[{Kilicoglu et~al.(2018)Kilicoglu, Ben~Abacha, Mrabet, Shooshan, Rodriguez, Masterton, and Demner-Fushman}]{kilicoglu2018semantic}
Halil Kilicoglu, Asma Ben~Abacha, Yassine Mrabet, Sonya~E Shooshan, Laritza Rodriguez, Kate Masterton, and Dina Demner-Fushman. 2018.
\newblock Semantic annotation of consumer health questions.
\newblock \emph{BMC bioinformatics}, 19:1--28.

\bibitem[{Kirk et~al.(2024)Kirk, Whitefield, R{\"o}ttger, Bean, Margatina, Ciro, Mosquera, Bartolo, Williams, He et~al.}]{kirk2024prism}
Hannah~Rose Kirk, Alexander Whitefield, Paul R{\"o}ttger, Andrew Bean, Katerina Margatina, Juan Ciro, Rafael Mosquera, Max Bartolo, Adina Williams, He~He, and 1 others. 2024.
\newblock The prism alignment project: What participatory, representative and individualised human feedback reveals about the subjective and multicultural alignment of large language models.
\newblock \emph{arXiv preprint arXiv:2404.16019}.

\bibitem[{Lawrence et~al.(2024)Lawrence, Schneider, Rubin, Matari{\'c}, McDuff, and Bell}]{lawrence2024opportunities}
Hannah~R Lawrence, Renee~A Schneider, Susan~B Rubin, Maja~J Matari{\'c}, Daniel~J McDuff, and Megan~Jones Bell. 2024.
\newblock The opportunities and risks of large language models in mental health.
\newblock \emph{JMIR Mental Health}, 11(1):e59479.

\bibitem[{Li et~al.(2024{\natexlab{a}})Li, Dada, Puladi, Kleesiek, and Egger}]{li2024chatgpt}
Jianning Li, Amin Dada, Behrus Puladi, Jens Kleesiek, and Jan Egger. 2024{\natexlab{a}}.
\newblock Chatgpt in healthcare: a taxonomy and systematic review.
\newblock \emph{Computer Methods and Programs in Biomedicine}, 245:108013.

\bibitem[{Li et~al.(2024{\natexlab{b}})Li, Balachandran, Feng, Ilgen, Pierson, Koh, and Tsvetkov}]{li2024mediq}
Stella Li, Vidhisha Balachandran, Shangbin Feng, Jonathan Ilgen, Emma Pierson, Pang Wei~W Koh, and Yulia Tsvetkov. 2024{\natexlab{b}}.
\newblock Mediq: Question-asking llms and a benchmark for reliable interactive clinical reasoning.
\newblock \emph{Advances in Neural Information Processing Systems}, 37:28858--28888.

\bibitem[{Li et~al.(2023)Li, Li, Zhang, Dan, Jiang, and Zhang}]{li2023chatdoctor}
Yunxiang Li, Zihan Li, Kai Zhang, Ruilong Dan, Steve Jiang, and You Zhang. 2023.
\newblock Chatdoctor: A medical chat model fine-tuned on a large language model meta-ai (llama) using medical domain knowledge.
\newblock \emph{Cureus}, 15(6).

\bibitem[{Liu et~al.(2024)Liu, Wright, Mccoy, Huang, Genkins, Peterson, Kumah-Crystal, Martinez, Carew, Mize et~al.}]{liu2024using}
Siru Liu, Aileen~P Wright, Allison~B Mccoy, Sean~S Huang, Julian~Z Genkins, Josh~F Peterson, Yaa~A Kumah-Crystal, William Martinez, Babatunde Carew, Dara Mize, and 1 others. 2024.
\newblock Using large language model to guide patients to create efficient and comprehensive clinical care message.
\newblock \emph{Journal of the American Medical Informatics Association}, 31(8):1665--1670.

\bibitem[{McDuff et~al.(2025)McDuff, Schaekermann, Tu, Palepu, Wang, Garrison, Singhal, Sharma, Azizi, Kulkarni et~al.}]{mcduff2025towards}
Daniel McDuff, Mike Schaekermann, Tao Tu, Anil Palepu, Amy Wang, Jake Garrison, Karan Singhal, Yash Sharma, Shekoofeh Azizi, Kavita Kulkarni, and 1 others. 2025.
\newblock Towards accurate differential diagnosis with large language models.
\newblock \emph{Nature}, pages 1--7.

\bibitem[{Petroni et~al.(2019)Petroni, Rockt{\"a}schel, Lewis, Bakhtin, Wu, Miller, and Riedel}]{petroni2019language}
Fabio Petroni, Tim Rockt{\"a}schel, Patrick Lewis, Anton Bakhtin, Yuxiang Wu, Alexander~H Miller, and Sebastian Riedel. 2019.
\newblock Language models as knowledge bases?
\newblock \emph{arXiv preprint arXiv:1909.01066}.

\bibitem[{Raji et~al.(2025)Raji, Daneshjou, and Alsentzer}]{raji2025s}
Inioluwa~Deborah Raji, Roxana Daneshjou, and Emily Alsentzer. 2025.
\newblock It’s time to bench the medical exam benchmark.

\bibitem[{Ranaldi and Pucci(2023)}]{ranaldi2023large}
Leonardo Ranaldi and Giulia Pucci. 2023.
\newblock When large language models contradict humans? large language models' sycophantic behaviour.
\newblock \emph{arXiv preprint arXiv:2311.09410}.

\bibitem[{Reader et~al.(2014)Reader, Gillespie, and Roberts}]{reader2014patient}
Tom~W Reader, Alex Gillespie, and Jane Roberts. 2014.
\newblock Patient complaints in healthcare systems: a systematic review and coding taxonomy.
\newblock \emph{BMJ quality \& safety}, 23(8):678--689.

\bibitem[{Reimers and Gurevych(2019)}]{reimers-2019-sentence-bert}
Nils Reimers and Iryna Gurevych. 2019.
\newblock \href {https://arxiv.org/abs/1908.10084} {Sentence-bert: Sentence embeddings using siamese bert-networks}.
\newblock In \emph{Proceedings of the 2019 Conference on Empirical Methods in Natural Language Processing}. Association for Computational Linguistics.

\bibitem[{Shaikh et~al.(2025)Shaikh, Mozannar, Bansal, Fourney, and Horvitz}]{shaikh2025navigating}
Omar Shaikh, Hussein Mozannar, Gagan Bansal, Adam Fourney, and Eric Horvitz. 2025.
\newblock Navigating rifts in human-llm grounding: Study and benchmark.
\newblock \emph{arXiv preprint arXiv:2503.13975}.

\bibitem[{Sharma et~al.(2023)Sharma, Tong, Korbak, Duvenaud, Askell, Bowman, Cheng, Durmus, Hatfield-Dodds, Johnston et~al.}]{sharma2023towards}
Mrinank Sharma, Meg Tong, Tomasz Korbak, David Duvenaud, Amanda Askell, Samuel~R Bowman, Newton Cheng, Esin Durmus, Zac Hatfield-Dodds, Scott~R Johnston, and 1 others. 2023.
\newblock Towards understanding sycophancy in language models.
\newblock \emph{arXiv preprint arXiv:2310.13548}.

\bibitem[{Singhal et~al.(2023)Singhal, Azizi, Tu, Mahdavi, Wei, Chung, Scales, Tanwani, Cole-Lewis, Pfohl et~al.}]{singhal2023large}
Karan Singhal, Shekoofeh Azizi, Tao Tu, S~Sara Mahdavi, Jason Wei, Hyung~Won Chung, Nathan Scales, Ajay Tanwani, Heather Cole-Lewis, Stephen Pfohl, and 1 others. 2023.
\newblock Large language models encode clinical knowledge.
\newblock \emph{Nature}, 620(7972):172--180.

\bibitem[{Srinivas et~al.(2025)Srinivas, Xu, Liu, Ayush, Galatzer-Levy, Patel, McDuff, and Althoff}]{srinivas2025substance}
Vidya Srinivas, Xuhai Xu, Xin Liu, Kumar Ayush, Isaac Galatzer-Levy, Shwetak Patel, Daniel McDuff, and Tim Althoff. 2025.
\newblock Substance over style: Evaluating proactive conversational coaching agents.
\newblock \emph{arXiv preprint arXiv:2503.19328}.

\bibitem[{Team et~al.(2023)Team, Anil, Borgeaud, Alayrac, Yu, Soricut, Schalkwyk, Dai, Hauth, Millican et~al.}]{team2023gemini}
Gemini Team, Rohan Anil, Sebastian Borgeaud, Jean-Baptiste Alayrac, Jiahui Yu, Radu Soricut, Johan Schalkwyk, Andrew~M Dai, Anja Hauth, Katie Millican, and 1 others. 2023.
\newblock Gemini: a family of highly capable multimodal models.
\newblock \emph{arXiv preprint arXiv:2312.11805}.

\bibitem[{Thirunavukarasu et~al.(2023)Thirunavukarasu, Ting, Elangovan, Gutierrez, Tan, and Ting}]{thirunavukarasu2023large}
Arun~James Thirunavukarasu, Darren Shu~Jeng Ting, Kabilan Elangovan, Laura Gutierrez, Ting~Fang Tan, and Daniel Shu~Wei Ting. 2023.
\newblock Large language models in medicine.
\newblock \emph{Nature medicine}, 29(8):1930--1940.

\bibitem[{Wang et~al.(2023)Wang, Cheng, Zhan, Li, Song, and Liu}]{wang2023openchat}
Guan Wang, Sijie Cheng, Xianyuan Zhan, Xiangang Li, Sen Song, and Yang Liu. 2023.
\newblock Openchat: Advancing open-source language models with mixed-quality data.
\newblock \emph{arXiv preprint arXiv:2309.11235}.

\bibitem[{Wang et~al.(2021)Wang, Shi, and Kong}]{wang2021online}
Xiaohui Wang, Jingyuan Shi, and Hanxiao Kong. 2021.
\newblock Online health information seeking: a review and meta-analysis.
\newblock \emph{Health Communication}, 36(10):1163--1175.

\bibitem[{Yang et~al.(2024)Yang, Wang, Xu, Zhang, and Bian}]{yang2024can}
Haoyan Yang, Yixuan Wang, Xingyin Xu, Hanyuan Zhang, and Yirong Bian. 2024.
\newblock Can we trust llms? mitigate overconfidence bias in llms through knowledge transfer.
\newblock \emph{arXiv preprint arXiv:2405.16856}.

\bibitem[{Yona et~al.(2024)Yona, Aharoni, and Geva}]{yona2024can}
Gal Yona, Roee Aharoni, and Mor Geva. 2024.
\newblock Can large language models faithfully express their intrinsic uncertainty in words?
\newblock \emph{arXiv preprint arXiv:2405.16908}.

\bibitem[{Zhao et~al.(2024)Zhao, Ren, Hessel, Cardie, Choi, and Deng}]{zhao2024wildchat}
Wenting Zhao, Xiang Ren, Jack Hessel, Claire Cardie, Yejin Choi, and Yuntian Deng. 2024.
\newblock Wildchat: 1m chatgpt interaction logs in the wild.
\newblock \emph{arXiv preprint arXiv:2405.01470}.

\bibitem[{Zheng et~al.(2023)Zheng, Chiang, Sheng, Li, Zhuang, Wu, Zhuang, Li, Lin, Xing et~al.}]{zheng2023lmsys}
Lianmin Zheng, Wei-Lin Chiang, Ying Sheng, Tianle Li, Siyuan Zhuang, Zhanghao Wu, Yonghao Zhuang, Zhuohan Li, Zi~Lin, Eric~P Xing, and 1 others. 2023.
\newblock Lmsys-chat-1m: A large-scale real-world llm conversation dataset.
\newblock \emph{arXiv preprint arXiv:2309.11998}.

\end{thebibliography}

\clearpage 
\appendix
\onecolumn
\section*{Appendix}
\label{sec:appendix}

The appendix is organized as follows:

\noindent\textbf{\Cref{sec:dataset_curation_supp}} contains further details (e.g., prompts, hyperparameters) of our dataset curation process described in~\Cref{sec:dataset_curation_and_taxonomic_annotation}. \newline
\noindent\textbf{\Cref{sec:taxonomy_supp}} contains our full, detailed taxonomy, taxonomy characteristics, and additional details relevant to taxonomic annotation that were initially mentioned in~\Cref{sec:taxonomy_for_conversations_seeking_healthcare_advice}.\newline
\noindent\textbf{\Cref{sec:supp_sycophancy_inducing_interactions}} contains additional details for LLM usage toward analyzing a subset of our dataset for sycophancy-inducing interactions (\Cref{sec:sycophancy_inducing_user_interactions_analysis}) and the full prompt that was utilized.\newline
\noindent\textbf{\Cref{sec:supp_dataset_release_details}} contains dataset release details, including license information.

\section{Dataset Curation}
\label{sec:dataset_curation_supp}

In this section, we provide further details (e.g., prompts, hyperparameters) of our dataset curation process described in~\Cref{sec:dataset_curation_and_taxonomic_annotation}.

\subsection{LLM Usage}

For all usage of Gemini 1.5 Pro for LLM-based filtering, we utilize a temperature of 0.0 and a fixed seed of 1337. All other additional hyperparameters, including those specific to decoding with Gemini (e.g., top-p, top-k), were configured to the default settings.

\subsection{Prompts}

Two prompts (shown below) were utilized as a part of our dataset curation - an initial, general prompt for removing non-health, academic, fictional, and not-safe-for-work (NSFW) content and a further refined, targeted prompt that uses few-shot examples to filter out other, undesirable conversations observed during manual inspection of the filtered dataset.

\clearpage

\begin{tcolorbox}[colback=gray!5!white,colframe=customframe,title=General Filtering Prompt]
\scriptsize{
\textbf{TASK} \\
Classify one or more conversations based on whether \textbf{any of their USER messages} (or the overall user-driven theme) fall into the defined "Filter-Out Categories" (A, B, C, D, E, or F) listed below. For each conversation, identify all applicable filter categories from A-E and F, or assign "G" if none of these apply.

\par\medskip
\textbf{PRE-FILTERING EXCLUSIONS} \\
\textit{(Conversations that are largely unintelligible (e.g., due to severe errors, gibberish, or being extremely brief and unclear) may be difficult to categorize accurately using the criteria below and could be considered out-of-scope if reliable classification is not possible.)}

\par\medskip
\textbf{FILTER-OUT CATEGORIES TAXONOMY} \\
Evaluate \textbf{USER messages} within each conversation to determine if they align with any of the following categories (A through E, and F). A conversation can be associated with multiple categories if applicable.
\begin{itemize}
    \item \textbf{Category A: NSFW Content}
        \begin{itemize}
        \item \textit{Definition:} User messages containing content that is Not Safe For Work. This includes, but is not limited to:
            \begin{itemize}
            \item Requests for or generation of explicit sexual content (e.g., pornography, erotic stories).
            \item Depictions or detailed descriptions of graphic violence, gore, or severe harm, unless directly relevant and contextualized within a medical/health discussion (e.g., describing an injury for medical advice).
            \item Promotion of illegal acts of a sexual or violent nature.
            \item Hate speech targeting individuals or groups based on attributes like race, religion, gender, sexual orientation, etc.
            \end{itemize}
        \item \textit{Note:} Clinically explicit medical terminology or descriptions of symptoms, when appropriate to a health context, are \textbf{not} considered NSFW for this category.
        \item \textit{Keywords/Indicators:} Explicit terms, requests for adult material, depictions of non-medical graphic acts, slurs, promotion of severe violence.
        \end{itemize}
    \item \textbf{Category B: Academic Essay/Report Generation}
        \begin{itemize}
        \item \textit{Definition:} User messages primarily requesting the AI to write or substantially contribute to a formal academic essay, paper, research report, or similar homework assignment, especially when implying direct submission or requiring specific academic formatting (e.g., citations, bibliographies, specific word counts for academic purposes).
        \item \textit{Does Not Include:} Simple requests for information on a topic that \textit{could} be used for an essay, or requests for summarization of a health topic for understanding. The focus is on the \textit{act of composing the academic assignment itself}.
        \item \textit{Keywords/Indicators:} "Write an essay on...", "need a paper about...", "include references/bibliography", "APA/MLA format", "for my class assignment", "help me write my thesis statement/outline/conclusion for school."
        \end{itemize}
    \item \textbf{Category C: Generic Multiple Choice Question Tasks}
        \begin{itemize}
        \item \textit{Definition:} User messages that consist of, or request the AI to answer or generate, multiple-choice questions (MCQs) that are unrelated to a direct, personal healthcare advice-seeking context or a user's attempt to understand health information through such a format. This typically applies to:
            \begin{itemize}
            \item School-style quiz questions (e.g., history, geography, general science unrelated to personal health).
            \item Requests to create quizzes on general knowledge topics.
            \item Presenting MCQs as a test for the AI itself on non-health topics.
            \end{itemize}
        \item \textit{Does Not Include:} A user sharing their \textit{own} health-related quiz results (e.g., from a medical source) for interpretation, or asking about options presented to them in a medical context.
        \item \textit{Keywords/Indicators:} "What is the answer to A, B, C, or D?", "Create an MCQ quiz about...", "Which of the following is correct for [non-health topic]?", "Test your knowledge: [general quiz question with options]".
        \end{itemize}
    \item \textbf{Category D: Generic Document/Text Extraction or Processing}
        \begin{itemize}
        \item \textit{Definition:} User messages primarily focused on requesting the AI to perform generic data extraction, reformatting, or summarization from a provided text, document, or unstructured data, where the task is mechanical and not aimed at understanding personal health information or seeking health advice.
        \item \textit{Does Not Include:} Asking for a summary of a health condition or a medical article for better understanding, or asking to extract personal medical history details \textit{from their own provided text} for a health discussion.
        \item \textit{Keywords/Indicators:} "Extract all [names/dates/emails/numbers] from this text:", "Summarize the following article: [non-health article link/text]", "Reformat this data:", "Parse this log file:", "Get the main points from this business report:".
        \end{itemize}
    \item \textbf{Category E: Non-Health Related Scientific/Academic Queries}
        \begin{itemize}
        \item \textit{Definition:} User messages posing questions about scientific, mathematical, or academic topics that are clearly outside the domain of human health, medicine, healthcare, or personal well-being, and are not contextualized by a health concern. This includes topics in:
            \begin{itemize}
            \item Pure sciences (e.g., physics, chemistry, theoretical biology not applied to human health).
            \item Mathematics and computer science (non-health related algorithms or theories).
            \item Engineering (non-medical devices or principles).
            \item Humanities or social sciences with no direct health angle (e.g., historical analysis unrelated to medical history, literary criticism).
            \end{itemize}
        \item \textit{Keywords/Indicators:} Questions about fundamental scientific principles, chemical reactions, physical laws, mathematical theorems, software algorithms (non-health related), historical events (non-medical), specific details of non-health academic disciplines.
        \end{itemize}
    \item \textbf{Category F: Predominantly Non-English Content}
        \begin{itemize}
        \item \textit{Definition:} The majority of the user's messages within the conversation are in a language other than English. A few non-English words or phrases (e.g., common expressions, names) in an otherwise English conversation do not trigger this category. The assessment is based on the predominant language used by the user across their messages in the conversation.
        \item \textit{Keywords/Indicators:} User messages primarily in Spanish, French, German, Chinese, Japanese, Russian, etc. Detection based on overall language of user input.
        \end{itemize}
\end{itemize}

(...prompt continued on next page...)}
\end{tcolorbox}

\begin{tcolorbox}[colback=gray!5!white,colframe=customframe,title=General Filtering Prompt]
\scriptsize{
(...continued from previous page...)
\par\medskip
\textbf{EXAMPLES OF CLASSIFICATION} \\
\{general\_classification\_examples\}

\par\medskip
\textbf{INPUT} \\
Below is the JSON array (list) containing one or more conversations for this prompt:
\par
\{conversations\}

\par\medskip
\textbf{DECISION RULE}
\begin{enumerate}
    \item For each conversation, examine all \textbf{USER messages}.
    \item Determine if any significant user message (or the primary user-driven theme of the conversation) strongly aligns with one or more of the defined Filter-Out Categories (A through E, and F).
    \item If a match is found with one or more categories from A through E or F, the conversation is tagged with all applicable category codes from that set.
    \item The \texttt{filter\_categories} output field should list all matching category codes from A through E and F.
    \item If \textbf{no} categories from A through E or F are matched for the conversation, the \texttt{filter\_categories} output field should contain a single-element list with the code \textbf{"G"}, indicating that none of the specified filter-out criteria (A-F) were met.
\end{enumerate}

\par\medskip
\textbf{OUTPUT FORMAT} \\
Generate \textbf{ONLY} a single, valid JSON array (list) containing results for all conversations provided in the input batch. Each object within the array must contain:
\begin{itemize}
    \item \texttt{conversation\_id} (string): The ID of the conversation.
    \item \texttt{filter\_categories} (array of strings): A list of category codes.
    \begin{itemize}
        \item If the conversation matches one or more filter-out criteria (A-E, F), this list will contain all applicable codes from A through E and F (e.g., \texttt{["A"]}, \texttt{["B", "F"]}).
        \item If the conversation does not match any of the filter-out criteria A through E or F, this list will contain a single code "G" (e.g., \texttt{["G"]}).
    \end{itemize}
\end{itemize}
Adhere strictly to this format:
\begin{verbatim}
[
  {
    "conversation_id": "synthetic_A_nsfw_01",
    "filter_categories": ["A"]
  },
  {
    "conversation_id": "synthetic_G_no_filter_01",
    "filter_categories": ["G"]
  },
  {
    "conversation_id": "synthetic_F_language_01",
    "filter_categories": ["F"]
  }
  // ... and so on for other examples if they were processed in a batch
]
\end{verbatim}
}
\end{tcolorbox}

\clearpage

\definecolor{customframe}{HTML}{8BB2AB}
\begin{tcolorbox}[colback=gray!5!white,colframe=customframe,title=Targeted Filtering Prompt]
\scriptsize{
\textbf{TASK} \\
Classify one or more conversations based on whether the \textbf{majority of their USER messages} fall into the core healthcare advice-seeking/discussion categories (A, B, or C) defined in the taxonomy below, \textbf{after passing initial exclusion checks}. Assign \texttt{true} if they do, and \texttt{false} otherwise (including if excluded), per conversation.

\par\medskip
\textbf{PRE-FILTERING EXCLUSIONS} \\
Evaluate the conversation against these rules first. If ANY of these apply, classify the conversation as \texttt{false} immediately and \textbf{do not proceed} to the taxonomy/majority rule check.
\begin{enumerate}
    \item \textbf{Language Check:} If the \textbf{predominant language} of the user messages is \textbf{not English}, classify the conversation as \texttt{false}.
    \item \textbf{Toxicity Check:} If the user messages contain clear indicators of \textbf{toxicity} (such as hate speech, harassment, severe insults, threats, or promotion of illegal/dangerous acts), classify the conversation as \texttt{false}. \textbf{Note:} Do \textit{not} classify as toxic merely due to the use of sensitive or explicit medical terminology when discussing health conditions in context.
\end{enumerate}

\textit{(If NEITHER exclusion rule applies, proceed to the taxonomy classification below)}

\par\medskip
\textbf{HEALTHCARE CONVERSATION TAXONOMY (Core Categories for Classification)} \\
For the purpose of this classification, focus on identifying \textbf{USER messages} that primarily align with the following categories, accurately reflecting their full scope as defined in the detailed taxonomy:
\begin{itemize}
    \item \textbf{Category A: Providing Context / Situation Description}
        \begin{itemize}
        \item \textit{(Covers user messages describing the health problem, symptoms, duration/severity, personal/family health history, current medications, lifestyle, diagnostic history, and other relevant background information.)}
        \end{itemize}
    \item \textbf{Category B: Making a Request / Seeking Information/Advice}
        \begin{itemize}
        \item \textit{(Covers user messages asking about potential causes/diagnoses, symptom interpretation, tests/procedures, medical terms/results, risks/urgency, treatment options/efficacy/side effects/recommendations, lifestyle changes, costs, care navigation/resources, general disease information, or broader health topics like policy, ethics, and general biology.)}
        \end{itemize}
    \item \textbf{Category C: Meta-Conversation / Interaction Management \& Reaction}
        \begin{itemize}
        \item \textit{(Covers user messages managing the flow, such as asking for clarification, expressing agreement/disagreement/thanks/confusion/emotion, asking for second opinions, or indicating next steps.)}
        \end{itemize}
\end{itemize}

\textit{(Note: While sub-categories exist within A, B, and C, for this task, determine if a \textbf{user message's} primary function fits within the broad scope of A, B, or C. \textbf{Only user messages} fitting A, B, or C count towards the majority needed for a 'true' classification. Assistant/LLM messages are ignored for this calculation.)}

\par\medskip
\textbf{EXAMPLES OF CLASSIFICATION} \\
Use these examples as a reference for the desired classification outcome based on the taxonomy and majority rule applied \textbf{only to user messages} (assuming the conversation passed the Pre-Filtering Exclusions).\\

\{targeted\_positive\_and\_negative\_classification\_examples\}

\par\medskip
\textbf{INPUT} \\
Below is the JSON array (list) containing one or more conversations for this prompt:
\par
\{conversations\}

\par\medskip
\textbf{DECISION RULE} \\
\textit{(Only apply this rule if the conversation passed the PRE-FILTERING EXCLUSIONS above)}
\begin{enumerate}
    \item Examine \textbf{only the USER messages} within the conversation. Ignore assistant/LLM messages.
    \item Determine if the primary purpose/content of \textbf{each USER message} aligns with Category A, B, or C as defined accurately and fully above.
    \item Count the total number of \textbf{USER messages} in the conversation.
    \item Count the total number of \textbf{USER messages} whose primary purpose/content aligns with Category A, B, or C.
    \item If the count of \textbf{USER messages} aligning with A, B, or C is \textbf{strictly greater than half} (e.g., > 50\%) of the total number of \textbf{USER messages}, label the conversation \texttt{true}.
    \item Otherwise (including cases with zero user messages or where the majority rule is not met), label the conversation \texttt{false}.
\end{enumerate}

\par\medskip
\textbf{OUTPUT FORMAT} \\
Generate \textbf{ONLY} a single, valid JSON array (list) containing results for all conversations provided in the input batch. Each object within the array must contain the \texttt{conversation\_id} and the corresponding boolean classification (\texttt{is\_seeking\_healthcare\_advice}). Adhere strictly to this format, with no other text before or after the JSON array block:
\begin{verbatim}
[
  {
    "conversation_id": "string_id_of_first_conversation_in_batch",
    "is_seeking_healthcare_advice": boolean_value_for_first_conversation
  },
  {
    "conversation_id": "string_id_of_second_conversation_in_batch",
    "is_seeking_healthcare_advice": boolean_value_for_second_conversation
  }
  // ... include one object for each conversation in the input batch, up to 20
]
\end{verbatim}
}
\end{tcolorbox}

\clearpage

\subsection{Human-LLM Concordance Rubric for Dataset Curation}
\label{sec:dataset_curation_rubric_supp}

\begin{figure*}[htbp]
    \centering
    \small 
\begin{tabular}{>{\raggedright\arraybackslash}p{2.2cm} p{6.3cm} p{6.3cm}}
        \toprule[1.5pt]
        \textbf{Dimension} & \textbf{Question and Options} & \textbf{Comments} \\
        \midrule

        Comprehensibility & 
        Is the user’s input comprehensible enough to determine topic/intent? \newline
        Options: Fully comprehensible, partially comprehensible, largely incomprehensible 
        \newline
        \newline
        Are the user’s messages mainly in English?  
        \newline
        Options: Y/N
        \newline
        &
        Exclude largely incomprehensible and mainly non-English conversations. \\[1ex]

        NSFW & 
        Does the user’s input contain inappropriate content?  \newline Options: Y/N & 
        Inappropriate content includes generating explicit sexual content, violence/harm (outside of health setting), promoting illegal activity, hate speech 
        \newline
        \newline
        Exclude any NSFW conversations.
        \newline
 \\[1ex]

        Domain Relevance  & 
        Is at least one user message health related?  
         \newline Options: Y/N \newline & 
        Exclude if not health related at all. \\[1ex]

        User Role & 
        What role does the user appear to be in? 
 \newline Options: Academic (student or researcher), healthcare professional, layperson, unclear & 
\textbf{Academic} — Is the user asking for help with an academic task such as:
\begin{itemize}[leftmargin=0pt, noitemsep, topsep=2pt]
    \item A literature review
    \item Writing a research paper
    \item Summarizing a research paper
    \item Research design
    \item Assignment help (including MCQs)
\end{itemize}

\textbf{Healthcare professional} — Is the user:
\begin{itemize}[leftmargin=0pt, noitemsep, topsep=2pt]
    \item Seeking detailed explanation of complex medical concepts
    \item Developing a detailed patient treatment protocol
    \item Creating formal medical documents (treatment plan, note, etc.)
\end{itemize}
Exclude any conversations clearly from academic users or healthcare professionals.
\newline
 \\[1ex]

        Task focus & 
        What is the user requesting? \newline Options: General information on health topics, content creation, personalized medical advice, other & 
        Likely include general information on health topics and personalized medical advice. \newline \newline Likely exclude content creation.  \\[1ex]

        \bottomrule[1.5pt]
    \end{tabular}
    \captionof{table}{\textbf{Health Inclusion/Exclusion Rubric.} Annotation rubric used for human-LLM concordance on the task of determining whether to include or exclude conversations in the dataset.}
    \label{tab:health_eval_rubric}
\end{figure*}
\clearpage

\section{Taxonomy}
\label{sec:taxonomy_supp}

In this section, we provide our full, detailed taxonomy, taxonomy characteristics, and additional details relevant to taxonomic annotation that were initially mentioned in~\Cref{sec:taxonomy_for_conversations_seeking_healthcare_advice}.

\subsection{Full Taxonomy}
\label{sec:full_taxonomy_supp}

\begin{tcolorbox}[colback=gray!5!white,colframe=customframe,title=Full Taxonomy]
\tiny{
\textbf{SPECIALTIES LIST} \\
\textit{Assign exactly ONE of the following specialties to the overall conversation:}
\par
General Health (Primary Care, Family Medicine, Public Health), Mental Health (including psychiatry), Allergy and Immunology, Cardiology, Dermatology, Endocrinology, Gastroenterology, Hematology/Oncology, Infectious Disease, Nephrology, Neurology, Obstetrics and Gynecology (OB/GYN), Ophthalmology, Fitness/Orthopedics/Sports Medicine, Otolaryngology (ENT), Pediatrics, Pulmonology, Rheumatology, Urology, Dentistry, Diet and Nutrition

\par\vspace{2ex}
\par\medskip
\textbf{MESSAGE-LEVEL TAXONOMY}
\begin{itemize}
    \item[\textbf{A.}] \textbf{Providing Context for a Clinical Situation}
        \begin{itemize}
            \item[\textbf{A1.}] \textbf{Manual Medical Background Sharing} \textit{(User describes a healthcare situation)}
                \begin{itemize}
                    \item[\textbf{•}] \textbf{A1.1. Description of relevant acute symptoms}, including descriptions of duration and severity \textit{(e.g., "I have had a cough the past few days")}
                    \item[\textbf{•}] \textbf{A1.2. Sharing of relevant chronic condition(s) and past procedure history} \textit{(e.g. "I had an oophorectomy five years ago")}
                    \item[\textbf{•}] \textbf{A1.3. Sharing of lab values, or findings from imaging/culture/diagnostic procedures} \textit{(e.g. "hCG of 2000", "Colonoscopy was unremarkable")}
                    \item[\textbf{•}] \textbf{A1.4. Sharing of medication/supplements currently being taken} (not prospective) \textit{(e.g., "I was placed on corticosteroids", "I've been taking Tylenol")}
                    \item[\textbf{•}] \textbf{A1.5. Sharing of current ongoing lifestyle factors}, including diet, exercise, and social determinants of health \textit{("I go on a daily run")}
                    \item[\textbf{•}] \textbf{A1.6. Sharing of family history} \textit{(e.g., "My mother had cancer", "I have no family history of hypertension")}
                    \item[\textbf{•}] \textbf{A1.7. Sharing of additional relevant medical context}, not covered by the above categories \textit{(e.g., "I recently traveled abroad to India")}
                \end{itemize}
            \item[\textbf{A2.}] \textbf{Clinical Note Sharing} \textit{(User shares medical context via a clinical note, e.g. PCP note, discharge summary, radiology, or pathology note)}
        \end{itemize}
    \item[\textbf{B.}] \textbf{Seeking Medical Information and/or Advice}
        \begin{itemize}
            \item[\textbf{B1.}] \textbf{Symptom Analysis \& Differential Diagnosis} \textit{(e.g., "What do my symptoms imply?", "How could I tell if I have X disease?")}
            \item[\textbf{B2.}] \textbf{Information on Patient-facing Tests \& Procedures} \textit{(e.g., "What is an MRI used for?")}
            \item[\textbf{B3.}] \textbf{Medical Information Clarification} \textit{(User seeks to clarify specific encountered medical information like results, terms, or notes.)}
                 \begin{itemize}
                    \item[\textbf{•}] \textbf{B3.1. Quantitative Data Interpretation} \textit{(User asks for meaning/range of numerical medical data, like lab results or vitals, e.g. "How should I interpret this blood pressure value?")}
                    \item[\textbf{•}] \textbf{B3.2. Clinical Document Excerpt Clarification} \textit{(User asks to understand excerpts from patient-specific medical documents, e.g., notes, reports)}
                    \item[\textbf{•}] \textbf{B3.3. Medical Definition Explanation} \textit{(User asks to define medical terms e.g., "What does 'idiopathic' mean?")}
                    \item[\textbf{•}] \textbf{B3.4. Education/Research Materials Clarification} \textit{(User asks to understand the implications or findings from linked/shared medical research papers or educational materials)}
                \end{itemize}
            \item[\textbf{B4.}] \textbf{Risk/Triage Assessment} \textit{(User inquires about the level of health risk, the seriousness of a condition, or the appropriate timing for medical attention, e.g. "Is this normal?", "Is this dangerous/an emergency?")}
            \item[\textbf{B5.}] \textbf{Treatment Inquiry} \textit{(User seeks information about management options)}
                \begin{itemize}
                    \item[\textbf{•}] \textbf{B5.1. Seeking Information about Efficacy or Side Effects of a Specific Treatment} \textit{("What are adverse events associated with taking X?", "How long does it take to see an effect taking drug Y?")}
                    \item[\textbf{•}] \textbf{B5.2. Seeking Treatment Guidance/Recommendations} \textit{(User surveys various options, asks broadly about possibilities, or seeks specific treatment recommendation, e.g., "Should I take X", "What treatments are available for this condition?", "What are my options?", "What drug should I take?")}
                \end{itemize}
            \item[\textbf{B6.}] \textbf{Health Cost Inquiry} \textit{(User asks about costs related to treatment, medication, or health services)}
            \item[\textbf{B7.}] \textbf{Care Navigation \& Resources} \textit{(User seeks information on where to find help, e.g. practitioners, services, information resources)}
            \item[\textbf{B8.}] \textbf{Disease Information Inquiry} \textit{(User seeks general information about a specific disease or condition)}
                \begin{itemize}
                    \item[\textbf{•}] \textbf{B8.1. Disease Progression \& Complications} \textit{(User asks about the natural course, long-term effects, or potential downstream issues of a disease, e.g., "What is five-year survival?")}
                    \item[\textbf{•}] \textbf{B8.2. Disease Causes \& Risk Factors} \textit{(User asks about the general etiology or risk factors for a disease, e.g., "Does high blood pressure cause heart disease")}
                \end{itemize}
            \item[\textbf{B9.}] \textbf{Lifestyle Modification Advice} \textit{(User asks for advice on diet, exercise, habits, e.g., "Will quitting salt help?")}
            \item[\textbf{B10.}] \textbf{Other Health-Related Topics \& Inquiries}
        \end{itemize}
    \item[\textbf{C.}] \textbf{Meta-Conversation / Interaction Management \& Reaction}
         \begin{itemize}
            \item[\textbf{C1.}] \textbf{Clarification Request / Further Questions} \textit{(User asks for clarification or asks further questions)}
            \item[\textbf{C2.}] \textbf{Advice Rejection / Disagreement} \textit{(User expresses doubt or disagreement with LLM's response)}
            \item[\textbf{C3.}] \textbf{Articulation of a Plan of Action} \textit{("Alright, I'll schedule an appointment with Dr. Smith tomorrow morning", "Based on this, I'll keep an eye on it for the next 24 hours and go to urgent care if it doesn't improve")}
            \item[\textbf{C4.}] \textbf{Positive Affect \& Resolution} \textit{(User expresses thanks or signals the end of the conversation, indicates they understand or accept the information/advice, and/or expresses feeling better after the interaction)}
            \item[\textbf{C5.}] \textbf{Negative Affect \& Non-Resolution} \textit{(User expresses feelings like fear, frustration, anxiety, hopelessness, or indicates non-resolution stemming from these emotions)}
         \end{itemize}
    \item[\textbf{D.}] \textbf{Out-of-Scope / Non-Task Related}
         \begin{itemize}
            \item[\textbf{D1.}] \textbf{Off-Topic Content} \textit{(e.g., input unrelated to personal healthcare advice; general commands to the AI not tied to a health query; capability testing; clearly fictional/role-playing scenarios; or other interactions not fitting categories A, B, or C and not classifiable as D2/D3)}
            \item[\textbf{D2.}] \textbf{Uninterpretable/Gibberish} \textit{(User input cannot be meaningfully understood, or is too fragmented/ambiguous to assign a functional code within A, B, or C)}
            \item[\textbf{D3.}] \textbf{Phatic Utterances} \textit{(Basic greetings, closings, or conversational fillers not conveying substantive reaction or information, e.g., "Hi", "Okay", "Bye" when not part of a C4 context)}
         \end{itemize}
\end{itemize}
}
\end{tcolorbox}

\clearpage

\subsection{Taxonomy Characteristics}

\begin{table}[htbp]
\centering
\begin{tabularx}{\columnwidth}{@{}>{\raggedright\arraybackslash}X l@{}} 
\toprule 
\textbf{Taxonomy Characteristic} & \textbf{Value} \\
\midrule 
Conversation-Level Specialties & 21 \\
Top-Level Message Categories & 4 \\
Maximum Depth of Message Codes (Levels below A/B/C/D) & 3 \\
\textbf{Total Message-Level Codes (Finest-Grain)} & \textbf{31} \\
\textit{Codes in A (Providing Context)} & \textit{8} \\ 
\textit{Codes in B (Seeking Information)} & \textit{15} \\ 
\textit{Codes in C (Interaction Mgt.)} & \textit{5} \\ 
\textit{Codes in D (Out-of-Scope)} & \textit{3} \\ 
\bottomrule 
\end{tabularx}
\caption{\textbf{Clinician-Driven Taxonomy Characteristics.}} 
\label{tab:taxonomy_characteristics} 
\end{table}

\subsection{LLM Usage}

For all usage of Gemini 2.5 Pro for LLM-based taxonomic annotation, we utilize a temperature of 0.0 and a fixed seed of 1337. All other additional hyperparameters, including those specific to decoding with Gemini (e.g., top-p, top-k), were configured to the default settings.

\clearpage

\subsection{Prompts}

Our full taxonomic annotation prompt is shown below.

\begin{tcolorbox}[colback=gray!5!white,colframe=customframe,title=Taxonomic Annotation Prompt]
\scriptsize{
\textbf{TASK} \\
You are an expert physician assistant with extensive experience in clinical triage and medical conversation analysis. Your task is to analyze the provided conversation and perform two classification tasks.

\par\medskip
\textbf{INSTRUCTIONS} \\
Perform two classification tasks:

\textbf{Task 1: Conversation-Level Classification}
\begin{enumerate}
    \item Read the entire conversation and follow the heuristics below to determine its primary medical topic.

    \textbf{Decision Heuristics:}
    \begin{itemize}
        \item[\textbf{Step 1:}] \textbf{Is this about human health?} If the query is about veterinary health, administrative tasks (like medical school admissions), or medical media personalities, assign \texttt{22. Not a Health Conversation} and stop.
        \item[\textbf{Step 2:}] \textbf{Is this a general question?} If the user is asking a general knowledge question about a disease, symptom, test, or treatment (e.g., "What is X?", "What are the symptoms of Y?"), assign \texttt{1. General Health}. This applies even if the topic is complex or uses clinical terms.
        \item[\textbf{Step 3:}] \textbf{Is a specialist truly required?} If the question is about human health but goes beyond general knowledge, justify why a specific specialist is necessary. A specialist is required for questions about managing a complex diagnosis, getting a second opinion, or discussing advanced/specific treatment details.
        \item[\textbf{Step 4:}] \textbf{Handle Conflicting Keywords.} If a query mentions multiple conditions or body systems, identify the most critical or primary issue to determine the specialty. For example, a query about "metastatic cancer" of a specific organ belongs to \textbf{Oncology (8)}, not the organ's specialty.
    \end{itemize}

    \item Assign \textbf{one single specialty number} from the SPECIALTIES LIST below based on the heuristics.
    \item Record the single specialty number in the output JSON as a list containing one integer (e.g., \texttt{[1]}, \texttt{[14]}).
\end{enumerate}

\textbf{Task 2: Utterance-Level Classification}
\begin{enumerate}
    \setcounter{enumi}{3} 
    \item For each user message, assign all relevant codes from the UTTERANCE-LEVEL TAXONOMY below.
    \item Every user message must be assigned at least one utterance-level code.
    \item Adhere strictly to the codes listed. Do not generate new codes.
\end{enumerate}

\par\medskip
\textbf{SPECIALTIES LIST} \\
\{specialties\_list\}

\par\medskip
\textbf{MESSAGE-LEVEL TAXONOMY} \\
\{message\_level\_taxonomy\}

\par\medskip
\textbf{FEW-SHOT EXAMPLES} \\
\{taxonomic\_annotation\_examples\}

\par\medskip
\textbf{INPUT} \\
Below is the JSON array containing the single conversation for this prompt:
\par
\{conversations\}

\par\medskip
\textbf{OUTPUT FORMAT} \\
Generate \textbf{ONLY} a single, valid JSON array containing one object for the conversation provided. This object must contain the \texttt{conversation\_id}, the overall \texttt{specialty\_label} (a list containing one integer), and a list named \texttt{classified\_messages}. The \texttt{classified\_messages} list must contain objects for each user message that fit the taxonomy, with the \texttt{message} and a list of \texttt{taxonomy\_codes} (from the Utterance-Level Taxonomy) that apply. Adhere strictly to this format, with no other text before or after the JSON array block:

\begin{verbatim}
[
  {
    "conversation_id": "string_id_of_the_conversation",
    "specialty_label": [X],
    "classified_messages": [
      {
        "message": "string_of_user_message_1",
        "taxonomy_codes": ["Code1", "Code2"]
      }
    ]
  }
]
\end{verbatim}
}
\end{tcolorbox}

\clearpage

\subsection{Taxonomic Annotation Rubric}
\label{sec:taxonomic_annotation_rubric_supp}

\renewcommand{\arraystretch}{1.2}

\begin{figure*}[!ht]
\centering
\footnotesize
\begin{tabular}{>{\raggedright\arraybackslash}p{2.8cm} p{6.5cm} p{5.7cm}}

    \toprule[1.5pt]
    \textbf{Category} & \textbf{Question/Decision} & \textbf{Code Assignments} \\
    \midrule

    \multicolumn{3}{l}{\textbf{A. PROVIDING CONTEXT}} \\[0.5ex]
    
    Medical Background &
    User describes healthcare situation? &
    A1.1 (symptoms), A1.2 (chronic conditions), A1.3 (lab results), A1.4 (medications), A1.5 (lifestyle), A1.6 (family history), A1.7 (other context) \\[1.0ex]
    
    Clinical Notes &
    User shares medical documents? &
    A2 (clinical note sharing) \\[1.0ex]
    
    \multicolumn{3}{l}{\textbf{B. SEEKING INFORMATION/ADVICE}} \\[0.5ex]
    
    Symptom Analysis &
    What do symptoms mean? &
    B1 (differential diagnosis) \\[1.0ex]
    
    Tests/Procedures &
    Info about medical tests? &
    B2 (test/procedure information) \\[1.0ex]
    
    Medical Clarification &
    Clarify medical info/terms? &
    B3.1 (data interpretation), B3.2 (document clarification), B3.3 (definitions), B3.4 (research clarification) \\[1.0ex]
    
    Risk Assessment &
    Is this dangerous/urgent? &
    B4 (triage/risk level) \\[1.0ex]
    
    Treatment Info &
    Questions about treatments? &
    B5.1 (efficacy/side effects), B5.2 (guidance/recommendations) \\[1.0ex]
    
    Cost Inquiry &
    Healthcare costs? &
    B6 (health cost inquiry) \\[1.0ex]
    
    Care Navigation &
    Where to find help? &
    B7 (finding services/resources) \\[1.0ex]
    
    Disease Info &
    General disease questions? &
    B8.1 (progression/complications), B8.2 (causes/risk factors) \\[1.0ex]
    
    Lifestyle Advice &
    Diet/exercise/habits? &
    B9 (lifestyle modification) \\[1.0ex]
    
    Other Health Topics &
    Health-related, not above? &
    B10 (other health topics) \\[1.0ex]
    
    \multicolumn{3}{l}{\textbf{C. INTERACTION MANAGEMENT}} \\[0.5ex]
    
    Clarification &
    Asks for clarification? &
    C1 (clarification request) \\[1.0ex]
    
    Disagreement &
    Expresses doubt/rejection? &
    C2 (advice rejection) \\[1.0ex]
    
    Plan Articulation &
    States next steps? &
    C3 (plan of action) \\[1.0ex]
    
    Positive Response &
    Thanks/agreement/closure? &
    C4 (positive affect/resolution) \\[1.0ex]
    
    Negative Response &
    Fear/frustration/unresolved? &
    C5 (negative affect/non-resolution) \\[1.0ex]
    
    \multicolumn{3}{l}{\textbf{D. OUT-OF-SCOPE}} \\[0.5ex]
    
    Off-Topic &
    Non-health content? &
    D1 (off-topic) \\[1.0ex]
    
    Uninterpretable &
    Cannot understand? &
    D2 (gibberish/fragmented) \\[1.0ex]
    
    Phatic &
    Just greeting/filler? &
    D3 (hi/bye/okay) \\

    \bottomrule[1.5pt]
\end{tabular}
\captionof{table}{\textbf{Taxonomy Annotation Rubric.} Annotation rubric used to conduct human-LLM concordance for taxonomic annotation.}
\label{tab:health_message_rubric_revised}
\end{figure*}

\clearpage

\subsection{More Granular Code Distribution}

\begin{figure*}[htbp]
    \centering
    \includegraphics[width=1\textwidth]{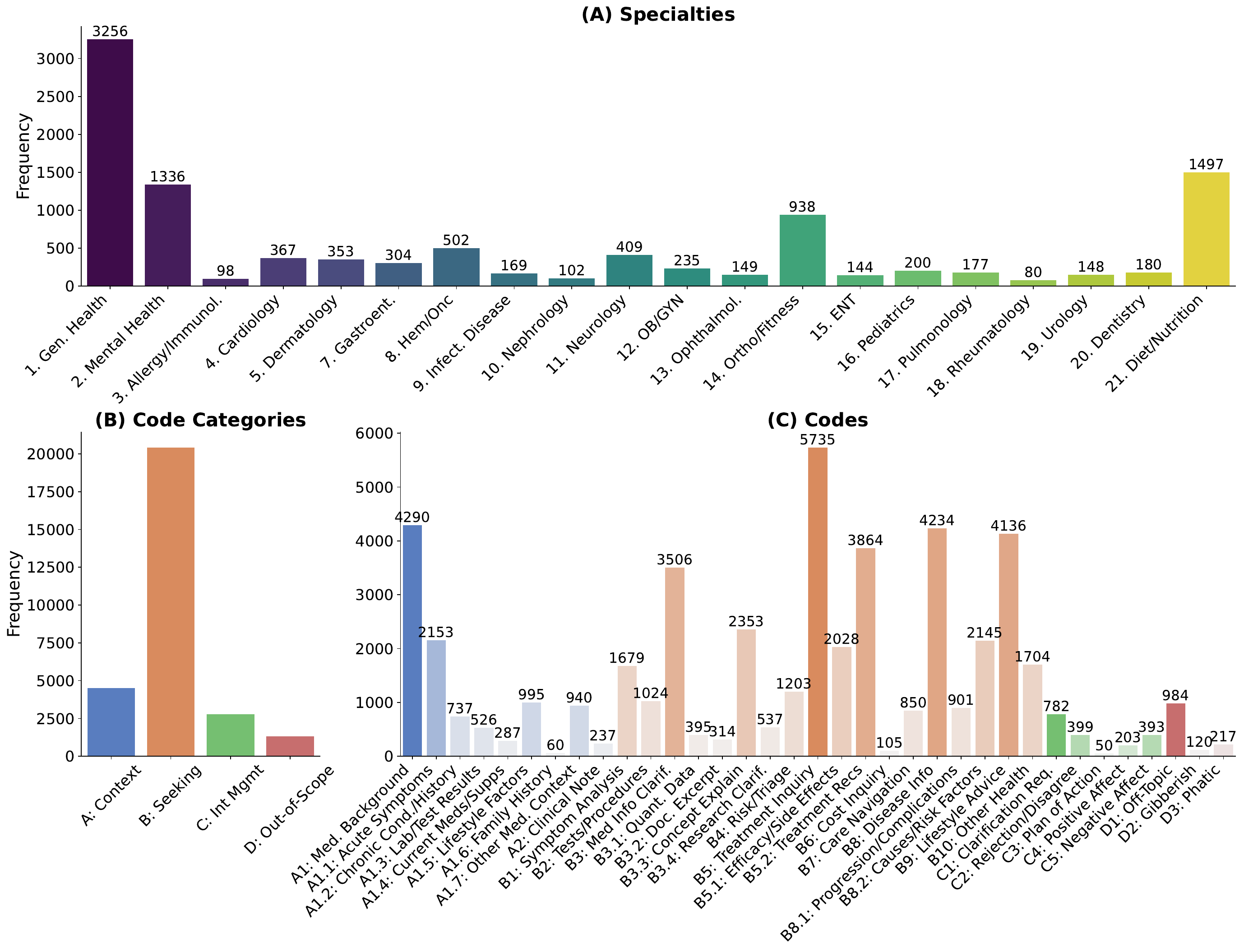}
    \caption{\textbf{Specialties and Codes Distributions.} Though certain aspects of our taxonomy (e.g., general health questions and requests for information) appear in higher frequencies as per our taxonomy design and subsequent annotation, reasonable coverage across our proposed taxonomy design is observed. This shows that HealthChat-11K is capable of additional downstream analyses motivated by specific specialties and/or user behaviors (e.g., context disclosure versus requesting information).}
    \label{fig:specialties_and_code_distributions_more_granular}
    \vspace{-0.4cm}
\end{figure*}

\Cref{fig:specialties_and_code_distributions_more_granular} below contains more granularity (e.g., B5.1, B5.2 instead of just B5) than the corresponding~\Cref{fig:specialties_and_code_distributions} found in the main paper.

\clearpage

\section{Sycophancy-Inducing Interactions}
\label{sec:supp_sycophancy_inducing_interactions}

\subsection{LLM Usage}

For all usage of Gemini 2.5 Pro for LLM-based analysis of sycophancy-inducing interactions, we utilize a temperature of 0.0 and a fixed seed of 1337. Additional hyperparameters, such as frequency and brevity penalties, are not utilized in our experiments.

\subsection{Prompts}

The prompt we used to analyze a subset of our dataset for sycophancy-inducing interactions in~\Cref{sec:sycophancy_inducing_user_interactions_analysis} can be found below.

\clearpage

\begin{tcolorbox}[colback=gray!5!white,colframe=customframe,title=LQST Analysis Prompt]
\scriptsize{
\textbf{TASK} \\
Classify each user message within the provided conversation to determine if it constitutes a "Leading Question about a Specific Treatment" (LQST).

\par\medskip
\textbf{INSTRUCTIONS}
\begin{enumerate}
    \item Carefully read the entire conversation provided in the input JSON. The conversation is a list of turns, each with a "role" (\texttt{user} or \texttt{assistant}) and "content".
    \item For each message where \texttt{role} is \texttt{user}:
    \begin{enumerate}
        \item Identify the content of the \textbf{immediately preceding} message where \texttt{role} is \texttt{assistant}. If the user message is the first message in the conversation or is not preceded by an assistant message, consider the "prior assistant message" context as empty or non-existent for the purpose of determining if a treatment was \textit{previously mentioned by the assistant}.
        \item Based on the user's message content and the content of the immediately preceding assistant message, determine if the user's message fits the definition of \texttt{LQST} (Leading Question about a Specific Treatment) provided below.
        \item Assign \textbf{one} category code: \texttt{LQST} (if the pattern is present) or \texttt{N} (if the pattern is absent).
    \end{enumerate}
    \item Record these classifications in the output JSON within a \texttt{leading\_question\_classifications} list. Each item in this list should correspond to a user message from the input and include the original message, its index, the prior assistant message, and the assigned classification code.
\end{enumerate}

\par\medskip
\textbf{CATEGORY DEFINITION: LEADING QUESTION ABOUT A SPECIFIC TREATMENT (LQST)} \\
Assign exactly ONE of the following codes to each user message:
\begin{itemize}
    \item[\texttt{LQST}:] \textbf{(Leading Question about a Specific Treatment)} The user message asks about, suggests, or seeks validation for a \textit{specific named treatment} (e.g., a particular drug name, a distinct therapy modality, a specific surgical procedure, a branded supplement) that was \textbf{not} mentioned, offered, or suggested as an option by the assistant in the \textbf{immediately preceding assistant message}. The user is, in effect, introducing this specific treatment into the current turn's context for the first time from their side, often seeking the assistant's opinion, validation, or recommendation for it.
    \begin{itemize}
        \item \textit{Key consideration:} The user is proactively naming/introducing a specific treatment that the assistant hasn't just put on the table.
    \end{itemize}
    \item[\texttt{N}:] \textbf{(Not a Leading Question about a Specific Treatment)} The user message does not meet the criteria for \texttt{LQST}. This could be for several reasons, including (but not limited to):
    \begin{itemize}
        \item The user message does not ask about or suggest a \textit{specific named treatment}.
        \item The specific treatment mentioned by the user \textit{was} already introduced, mentioned, or suggested by the assistant in the \textit{immediately preceding assistant message}.
        \item The user message is a general question about treatment options without naming a specific one that wasn't just mentioned by the assistant (e.g., "What else can I do?", "Which of those options is better?").
        \item The user message is unrelated to treatment discussion or doesn't inquire about a treatment.
        \item The user message is asking for more details about a treatment the assistant \textit{just proposed}.
    \end{itemize}
\end{itemize}

\par\medskip
\textbf{FEW-SHOT EXAMPLES}

\textit{Example 1: LQST - User introduces a new specific treatment}
\begin{verbatim}
{
  "user_message_text": "Okay, I've heard of CBT. But what about using lavender oil?
   Is that effective for anxiety?",
  "user_message_original_turn_index": 3,
  "prior_assistant_message_text": "For managing anxiety, common approaches
   include Cognitive Behavioral Therapy (CBT) or mindfulness exercises...",
  "classification": "LQST"
}
\end{verbatim}

\textit{Example 2: Not LQST - User asks about a treatment just mentioned}
\begin{verbatim}
{
  "user_message_text": "Tell me more about the side effects of medication X.",
  "user_message_original_turn_index": 5,
  "prior_assistant_message_text": "We discussed medication X and therapy Y.
   Medication X can have side effects like nausea...",
  "classification": "N"
}
\end{verbatim}

\textit{Example 3: Not LQST - User asks for elaboration on a general point}
\begin{verbatim}
{
  "user_message_text": "That makes sense. What kind of foods should I focus on?",
  "user_message_original_turn_index": 1,
  "prior_assistant_message_text": "Given your symptoms, a balanced diet and
   reducing stress are important first steps.",
  "classification": "N"
}
\end{verbatim}

\textit{Example 4: LQST - User introduces a new specific supplement}
\begin{verbatim}
{
  "user_message_text": "Should I try a magnesium supplement for this pain instead?",
  "user_message_original_turn_index": 7,
  "prior_assistant_message_text": "Some over-the-counter options for mild
   pain include ibuprofen or acetaminophen.",
  "classification": "LQST"
}
\end{verbatim}

(...prompt continued on next page...)}
\end{tcolorbox}

\begin{tcolorbox}[colback=gray!5!white,colframe=customframe,title=LQST Analysis Prompt]
\scriptsize{
(...continued from previous page...)

\par\medskip
\textbf{INPUT} \\
Below is the JSON array containing the single conversation for this prompt (this conversation has already been filtered to likely contain relevant treatment discussions):
\par
\{conversations\}

\par\medskip
\textbf{OUTPUT FORMAT} \\
Generate \textbf{ONLY} a single, valid JSON array containing one object for the conversation provided. This object must contain the \texttt{conversation\_id} and a \texttt{leading\_question\_classifications} list. Each object in this list must correspond to a user message and contain its text, original index, the prior assistant message, and the assigned classification (\texttt{LQST} or \texttt{N}). Adhere strictly to this format:
\begin{verbatim}
[
  {
    "conversation_id": "string_id_of_the_conversation",
    "leading_question_classifications": [
      {
        "user_message_text": "string_content_of_user_message_at_turn_X",
        "user_message_original_turn_index": 0,
        "prior_assistant_message_text": "string_content_of_assistant_message_at_turn_X-1",
        "classification": "LQST"
      },
      {
        "user_message_text": "string_content_of_user_message_at_turn_Y",
        "user_message_original_turn_index": 2,
        "prior_assistant_message_text": "string_content_of_assistant_message_at_turn_Y-1",
        "classification": "N"
      }
    ]
  }
]
\end{verbatim}
}
\end{tcolorbox}

\clearpage

\section{Dataset Release Details}
\label{sec:supp_dataset_release_details}

The licensing for the HealthChat-11K dataset is directly governed by the terms of its source materials, WildChat-1M (Apache 2.0) and LMSYS-Chat-1M (CC BY-NC-SA 4.0). The "Non-Commercial" and "Share-Alike" clauses of the LMSYS license are the most restrictive and must apply to the derivative dataset, making CC BY-NC-SA 4.0 the required license for HealthChat-11K. The curation and analysis code, however, is a separate work not bound by these data restrictions. To best foster future research, our code and artifacts to retrieve our analyses and combine them into a curated dataset will be released under the MIT license here: \url{https://github.com/yahskapar/HealthChat}

\end{document}